# A self-supervised text-vision framework for automated brain abnormality detection

David A. Wood, Emily Guilhem, Sina Kafiabadi, Ayisha Al Busaidi, Kishan Dissanayake, Ahmed Hammam, Nina Mansoor, Matthew Townend, Siddharth Agarwal, Yiran Wei, Asif Mazumder, Gareth J. Barker, Peter Sasieni, Sebastien Ourselin, James H. Cole, Thomas C. Booth

## Abstract

Artificial neural networks trained on large, expert-labelled datasets are considered state-of-the-art for a range of medical image recognition tasks. However, categorically labelled datasets are time-consuming to generate and constrain classification to a pre-defined, fixed set of classes. For neuroradiological applications in particular, this represents a barrier to clinical adoption.

To address these challenges, we present a self-supervised text-vision framework that learns to detect clinically relevant abnormalities in brain MRI scans by directly leveraging the rich information contained in accompanying free-text neuroradiology reports. Our training approach consisted of two-steps. First, a dedicated neuroradiological language model - NeuroBERT - was trained to generate fixed-dimensional vector representations of neuroradiology reports (N = 50,523) via domain-specific self-supervised learning tasks. Next, convolutional neural networks (one per MRI sequence) learnt to map individual brain scans to their corresponding text vector representations by optimising a mean square error loss.

Once trained, our text-vision framework can be used to detect abnormalities in unreported brain MRI examinations by scoring scans against suitable query sentences (e.g., 'there is an acute stroke', 'there is hydrocephalus' etc.), enabling a range of classification-based applications including automated triage. Potentially, our framework could also serve as a clinical decision support tool, not only by suggesting findings to radiologists and detecting errors in provisional reports, but also by retrieving and displaying examples of pathologies from historical examinations that could be relevant to the current case based on textual descriptors.

# 1. Introduction

Magnetic resonance imaging (MRI) plays a key role in the diagnosis and management of a range of neurological conditions (Atlas, 2009). However, the growing demand for brain MRI examinations, along with a global shortage of radiologists, is taking its toll on healthcare systems. Increasingly, radiologists are unable to fulfill their reporting requirements within contracted hours, leading to substantial reporting delays (NHS, 2021)(Wood et al., 2021). Concerns about fatigue-related diagnostic errors are also mounting as radiologists become increasingly overworked (Vosshenrich et al., 2021). Ultimately, reporting delays and errors lead to delays in treatment; for many abnormalities, this results in poorer patient outcomes and inflated healthcare costs (Adams et al., 2005).

Potentially, artificial intelligence (AI) could be used to relieve some of the pressure on radiology departments, for example by supporting real-time triaging of examinations (Annarumma et al., 2019)(Yala et al., 2019)(Wood et al., 2022)(Verburg et al., 2022)(Agarwal et al., 2023)(Booth et al., 2023)(Agarwal et al., 2023) or assisting radiologists to reduce errors in radiology reports. To date, efforts in this direction have largely relied on deep learning models trained on expert-labelled datasets (Gulshan, 2016))(Titano et al., 2018)(De Fauw et al., 2018)(Ardila et al., 2019)(McKinney et al., 2020)(Wood et al., 2019)(Wood et al., 2022)(Din et al., 2023)(Chelliah et al., 2024). However, there are key limitations to this approach. First, the growing pressure on clinical services has made it increasingly difficult to justify using radiologists' time to manually annotate images for research purposes; obtaining large, clinically representative training datasets therefore represents a bottleneck to model development (Wood et al., 2020)(Benger et al., 2023)(Wood et al., 2024). Second, the use of categorically labelled datasets in conjunction with supervised learning methods inherently restricts classification to a pre-defined, fixed set of classes. As such, whenever a new classification task emerges, additional labelled training examples are needed. This poses a considerable problem for neuroradiological applications, where the dynamic nature of clinical demands constantly alters the landscape of automation possibilities. For example, the class of 'tumours' may become insufficient for a detection task when there is a new demand for a particular type of tumour; additional labelling of the particular type of tumour is required (Louis et al., 2021).

These issues, among others, have led to a growing interest in multi-modal (e.g., text-vision) self-supervised methods which enable computer vision models to learn directly from free-text radiology reports (Zhang et al., 2022)(Boecking et al., 2022)(Bannur et al., 2023). Radiology reports represent promising training data since they i) contain detailed descriptions and impressions of all image findings observed by expert radiologists; and ii) are typically stored alongside imaging data on hospital picture archiving and communication systems (PACS) and so are relatively easy to obtain. To date, however, the application of self-supervised methods has largely been limited to image recognition tasks involving chest radiographs - due in part to the availability of open-access, paired image-text datasets such as MIMIC Chest X-ray (MIMIC-CXR) (Johnson et al., 2019)(Agarwal et al., 2024). To our knowledge there has been no previous demonstration of text-vision models for either brain abnormality detection or for the highly complex modality of MRI (Wood et al., 2022).

Here, we present a self-supervised text-vision framework which learns to detect clinically relevant abnormalities from unlabelled hospital brain MRI scans. Our two-step training approach proceeded as follows. First, a dedicated neuroradiological language model - NeuroBERT - was trained to generate fixed-dimensional vector representations of neuroradiology reports via domain-specific self-supervised learning tasks. Next, convolutional neural networks (CNN) - one per MRI sequence type, covering the full range of sequences performed during routine examinations - learnt to map individual brain scans to their corresponding text vector representations by optimising a mean square error (MSE) loss. Once trained, our text-vision framework can be used to detect abnormalities in unreported brain MRI examinations by scoring scans against suitable query sentences (e.g., 'this is a normal study', or 'there is an acute stroke' etc.), opening a range of classification-based applications including automated triage (Fig. 1), diagnosis, and treatment response assessment. Potentially, our framework could also operate as a clinical decision support tool by suggesting findings to radiologists, detecting errors in provisional reports, and retrieving and displaying examples of pathologies from historical examinations that could be relevant to the current case based on textual descriptors.

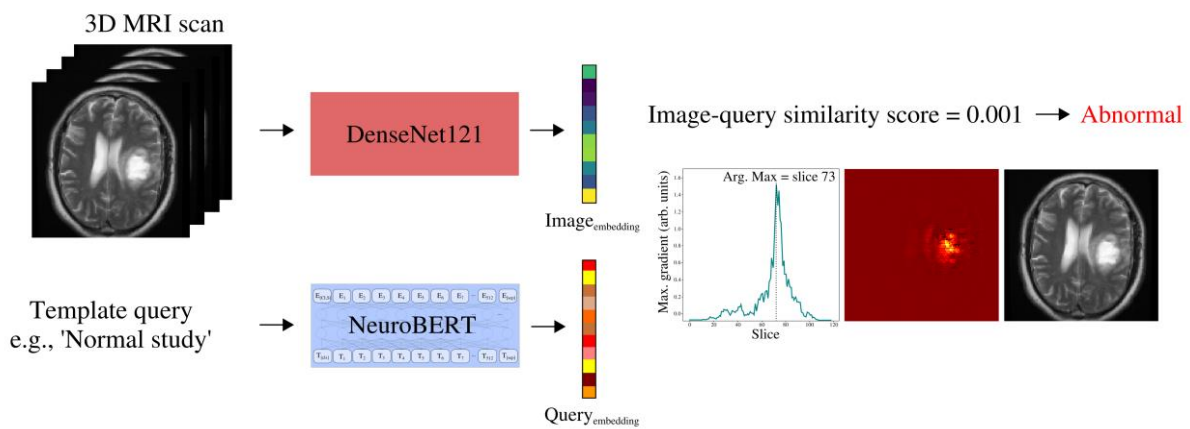

**Figure 1:** *Overview of our text-vision brain abnormality detection framework. A CNN-based image encoder learns to map 3D brain MRI scans to 'image embeddings' in a semantically meaningful vector space. Abnormality detection is then achieved by treating the problem as*

*a text-image similarity task and scoring scans against suitable query sentences. In this example, a scan containing a large extra-axial mass is scored against the sentence 'normal study'; the low text-image similarity score indicates that this scan is unlikely to be normal, and the saliency map helps to highlight where the potential abnormality is located. Unlike traditional supervised models which are restricted to categorising images into a pre-defined, fixed set of classes, our text-vision framework has the flexibility to generate similarity scores for arbitrary image findings. For example, the scan could just as easily be scored against more specific input sentences (e.g., 'there is a glioma', 'there is an acute stroke', 'there is a pineal cyst' etc.) to help identify the nature of the suspected abnormality (see section 4.2).*

## 2. Related work

Our work builds on previous studies which have sought to train computer vision models via natural language supervision, both in general and medical image settings. An important early work was that of Frome and colleagues (Frome et al, 2013) who introduced DeViSE, a deep visual-semantic embedding model. DeViSE included two components: i) the then state-of-the-art 'word2vec' language model (Mikolov et al., 2013); and ii) a CNN-based image encoder. Following initial pre-training on ImageNet, the CNN was fine-tuned to predict the vector embedding of the label text for each image. Crucially, semantically and visually similar ImageNet classes (e.g., 'great white shark', 'tiger shark', 'hammerhead shark', etc.) have similar word2vec embeddings, enabling the CNN to leverage information about class relationships during training. In contrast, the common approach of formulating ImageNet as a discrete N-way classification task (i.e., using one-hot encoded labels) treats all classes as disconnected and unrelated.

The next important study was that of Zhang and colleagues (Zhang et al., 2020) who introduced ConViRT, a framework for learning medical image representations from raw text. Like DeViSE, ConViRT consists of two components - a text encoder and an image encoder. Unlike DeViSE, however, ConViRT relies on a contrastive learning loss which seeks to maximise the agreement between image and text representations for true image-text pairs and minimize the agreement for randomly sampled pairs. Originally, ConViRT was presented as a pre-training framework; downstream classification required that a linear output layer be added, and supervised fine-tuning be performed using a labelled dataset. Nonetheless, ConViRT demonstrated impressive label efficiency, achieving state-of-the-art performance on a range of chest radiograph recognition tasks using an order of magnitude less labelled data than competing supervised learning approaches.

Shortly after the development of ConViRT, Radford and colleagues (Radford et al., 2021) introduced Contrastive Language-Image Pretraining (CLIP), a framework for contrastive text-image modelling in the general image setting. The architecture of CLIP is identical to ConViRT, but instead of using paired medical images and radiology reports, CLIP was trained on a large dataset of image-caption pairs (~400 million) collected from the internet. Importantly for our study, the authors demonstrate that CLIP can classify images without supervised fine-tuning by recasting the problem as a text-image similarity task and scoring images against query sentences (e.g., 'this is a photo of a dog', 'this is a photo of a cat' etc.).

The next relevant study was that of Huang and colleagues (Huang et al., 2021) who introduced GLoRIA, a multimodal global-local contrastive learning framework for label-efficient medical image recognition. Like ConViRT, GLoRIA includes a global contrastive loss i.e., one which seeks to maximise the agreement between representations for entire images and reports; however, GLoRIA also includes a local loss which contrasts attention-weighted image sub-regions with individual words in reports. An advantage of this approach is that the resulting attention heatmaps can be used as a form of model explainability. However, a limitation is that words describing the absence of a finding (e.g., 'there are no areas of restricted diffusion'), as well as words describing diffuse pathology, do not map directly to specific localised image sub-regions (e.g., 'there is generalised volume loss').

The final study relevant to our work is that of Boecking and colleagues (Boecking et al., 2022) who introduced BioViL, a text-vision model optimised for paired biomedical data. BioViL is similar to GLoRIA in that it includes both a global and local contrastive loss function; however, BioViL also includes domain-specific language pre-training tasks which improves the model's ability to capture the semantics of radiology reports. Most pertinent to our work is the inclusion of a radiology report section matching task, formulated to identify 'Findings' sections (typically a complete description of the images) and 'Summary' sections of the same study, and favour these over randomly sampled pairs from different studies.

## 3. Methods and materials

### 3.1 Self-supervised text-vision framework

Our self-supervised text-vision framework consists of a dedicated neuroradiological language model - which we call NeuroBERT - and CNN-based image encoders. We discuss these separately below.

#### 3.1.1 NeuroBERT

NeuroBERT plays two key roles in our text-vision framework: i) it generated fixed-dimensional vector representations, or 'embeddings', of hospital brain neuroradiology reports which formed the training signal for our computer vision models; and ii) at test-time it embeds query sentences to facilitate abnormality detection in unseen scans (Fig. 1).

Architecturally, NeuroBERT is identical to BERT, a state-of-the-art general-purpose transformer language model (Devlin et al., 2018). However, important modifications to BERT's training procedure were required to optimise NeuroBERT for neuroradiology reports; we discuss these separately below.

### 3.1.1.1 Custom neuroradiological vocabulary

Transformer-based language models are unable to directly 'read' raw text. Instead, they rely on tokenization, a process in which text is first split into smaller units (i.e., words, characters etc.) and then converted to machine readable IDs. To keep the vocabulary size manageable, BERT uses an algorithm called WordPiece tokenization (Wu et al., 2016) which assigns unique IDs to common words but breaks rarer words up into sub-word pieces. However, this poses a problem when working with complex neuroradiological text, since many words which are commonly found in neuroradiology reports - but not in everyday language - are broken up into pieces which have no radiological relevance (e.g., BERT breaks 'haemorrhage' into five pieces: 'ha', 'em', 'or', 'rh', and 'age'). Clearly, this is sub-optimal for downstream learning.

To overcome this, we trained a custom WordPiece vocabulary of 10,000 words using our corpus of unlabelled hospital neuroradiology reports (see section 3.2). Using this dedicated neuroradiological vocabulary, NeuroBERT produces far fewer word breakdowns compared with general purpose vocabularies (Table A1 in Appendix A). We make this custom vocabulary and tokenizer available at https://github.com/MIDIconsortium/NeuroBERT.

### 3.1.1.2 Masked language modelling without next sentence prediction

The original BERT model was trained via two self-supervised learning tasks: next sentence prediction (NSP) and masked language modelling (MLM). In the context of complex neuroradiology reports, however, the frequent occurrence of adjacent but unrelated sentences (e.g., 'There is a mild degree of right hippocampal volume loss. Incidental note is made of a small left-sided arachnoid cyst.') makes NSP an inappropriate training objective. We therefore omit this task, and initially train NeuroBERT via MLM only (Fig. 2).

Our approach to MLM largely follows the original BERT paper (Devlin et al., 2018). We randomly replaced 15% of the tokens in our neuroradiology report corpus with the special <mask> token. Short input sequences were then fed into NeuroBERT and transformed, via a series of stacked self-attention layers, into 768-dimensional contextualised word embeddings[1]. We then passed the <mask> embeddings ($E_2$ and $E_7$ in the example shown in

---

[1] The default dimensionality of BERT embeddings is 768; we elected to keep this dimensionality for NeuroBERT embeddings in order to ensure compatibility with existing BERT-based model architectures.

Fig. 2) through a feed-forward neural network with a softmax layer to generate a probability distribution over our custom vocabulary. Finally, the cross-entropy loss between NeuroBERT's predictions and the ground-truth mask labels was computed, and the model weights were updated by gradient backpropagation.

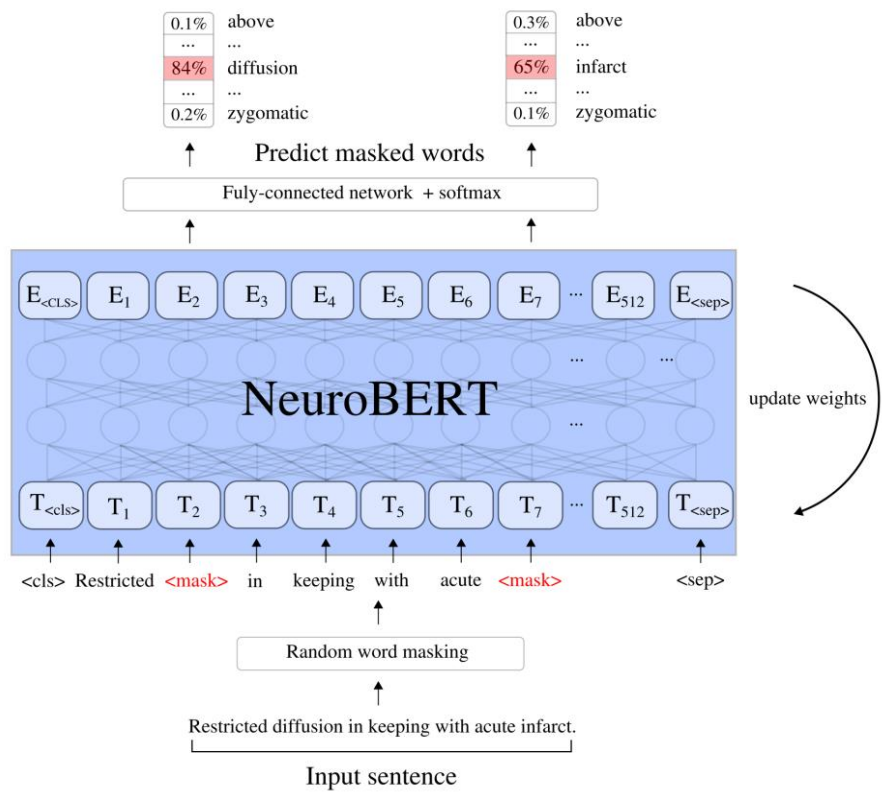

*Figure 2: Overview of our masked language modelling (MLM) training procedure. MLM is a self-supervised learning task which involves predicting the identity of randomly masked tokens using the context provided by surrounding words. In this example, the words 'diffusion' and 'infarct' have been masked from the sentence 'restricted diffusion in keeping with acute infarct' and must be predicted. Following the original BERT training procedure, we passed the final hidden-layer embeddings of the <mask> tokens ($E_2$ and $E_7$ in this example) through a feed-forward neural network with a softmax layer to generate a probability distribution over our custom vocabulary. The cross-entropy loss between NeuroBERT's predictions and the ground-truth mask labels was computed, and the model weights were updated by gradient backpropagation.*

### 3.1.1.3 Radiology report section matching

MLM is considered state-of-the-art for encoder-based language model pre-training and leads to the generation of rich contextualised word embeddings. For our purposes, however, word-level representation is insufficient; we require fixed-dimensional vector representations which capture the underlying semantics of entire neuroradiology reports. To this end, and inspired by (Boecking et al., 2022), we additionally trained NeuroBERT to perform radiology section matching (RSM), a domain-specific self-supervised learning task formulated to encourage the generation of similar vector representations for 'Findings' and 'Summary' sections from the same report.

The motivation behind RSM is straightforward. Findings sections typically contain a description of all abnormalities which have been observed, as well as those which have been ruled out (e.g., '*Findings: The brain volume is normal for age. There are no intra or extra-axial mass lesions. There are no areas of haemorrhage. There is a focus of restricted diffusion in the left MCA region in keeping with an acute infarct*'). In contrast, Summary sections provide only a concise recapitulation of the salient image findings to ensure clarity for the referring non-specialist clinician (e.g., the Summary section for the above example might read '*Summary: Restricted diffusion in keeping with an acute left MCA infarct*'). Therefore, to effectively perform this matching task NeuroBERT must learn to identify and prioritise the most relevant information in neuroradiology reports, and preferentially include this information in its embeddings.

We adopted the Siamese network architecture of SentenceBERT (Reimers and Gurevych, 2019) - a state-of-the-art framework for semantic similarity modelling – for our RSM training procedure and created two copies of NeuroBERT with tied (i.e., constrained to be identical) weights (Fig. 3). Findings and Summary sections from either the same report (true-pairs) or from separate, randomly sampled reports (false-pairs) were tokenized and fed into the network. Mean pooling was then applied to generate section-level embeddings (i.e., fixed-length vector representations of Findings and Summary sections), and a cosine similarity score between -1 and 1 was calculated. We used the MSE between the similarity score and the ground-truth label (1 for true-pairs, 0 for false-pairs) as the training objective and applied gradient backpropagation to update the model weights.

Following RSM training, NeuroBERT takes free-text neuroradiology reports as input and returns concise 'vector summaries' which are then used as latent, high-dimensional 'labels' for computer vision training (see section 3.1.2)[2]. We make our trained NeuroBERT model available to other researchers at https://huggingface.co/sentence-transformers/NeuroBERT.

---

[2] We observed no significant difference in downstream performance between passing the Findings section, the Summary section, or the entire report as input to NeuroBERT when generating report embeddings. For simplicity, we elected to pass the entire report when generating embeddings for computer vision training.

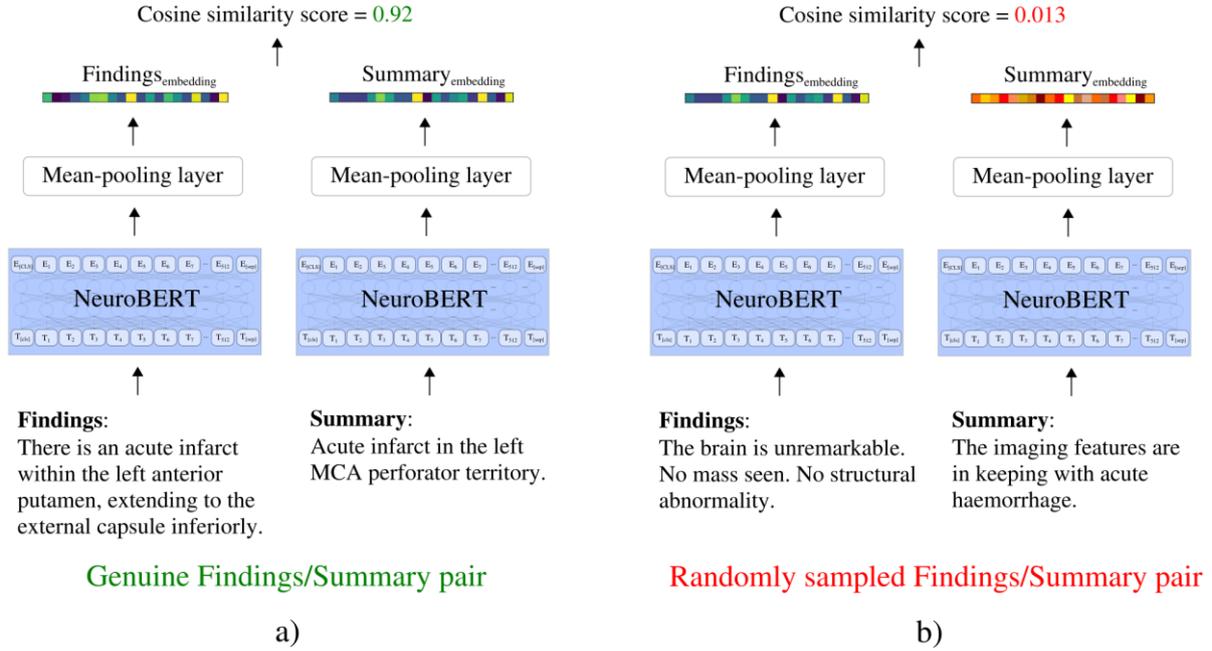

**Figure 3:** *Overview of our radiology section matching (RSM) procedure. RSM is a domain-specific self-supervised task formulated to encourage the generation of similar vector representations for 'Findings' and 'Summary' sections from the same report (a), and discourage similar representations for sections from separate, randomly sampled reports (b). We adopted a Siamese network architecture (i.e., two copies of NeuroBERT with tied weights) and applied mean pooling to generate section embeddings. The training objective was the MSE between the cosine similarity score of the embeddings and the ground-truth label (i.e., 1 for true-pairs, 0 for false pairs), and gradient backpropagation was used to update the model weights.*

### 3.1.2 CNN Image encoder

The purpose of our CNN image encoders is to map 3D brain MRI scans to vectors in the same, semantically meaningful, vector space as that used by NeuroBERT. Unlike other medical imaging modalities, however, MRI examinations typically comprise multiple image 'sequences' (e.g., $T_1$-weighted images, $T_2$-weighted images, diffusion-weighted images etc.). Therefore, we developed separate CNN models to cover the sequences most commonly performed during routine clinical imaging (see section 3.2).

Each single-sequence model is based on a modified version of the popular DenseNet121 architecture (Huang et al., 2017). We kept the four 'dense blocks' and 'transition layers' from the original network but dropped the final linear layer and softmax operation. Instead, we applied global average pooling to the output of the 4[th] dense block, concatenated the result with the patient's age (to capture age-dependent abnormalities such as excessive volume loss etc.), and then feed this into a single-layer feedforward neural network to generate 768-dimensional age-conditional 'image embeddings'.

During the training process, we sample batches of paired 3D brain MRI scans and corresponding neuroradiology reports from our training dataset (see section 3.2). These batches are then fed into the relevant sequence image encoder and NeuroBERT, respectively (Fig. 4). Our training objective is defined by a simple element-wise mean square error (MSE) loss between the image and report embeddings, and gradient backpropagation is applied to update the weights of the image encoder, keeping NeuroBERT's parameters constant.

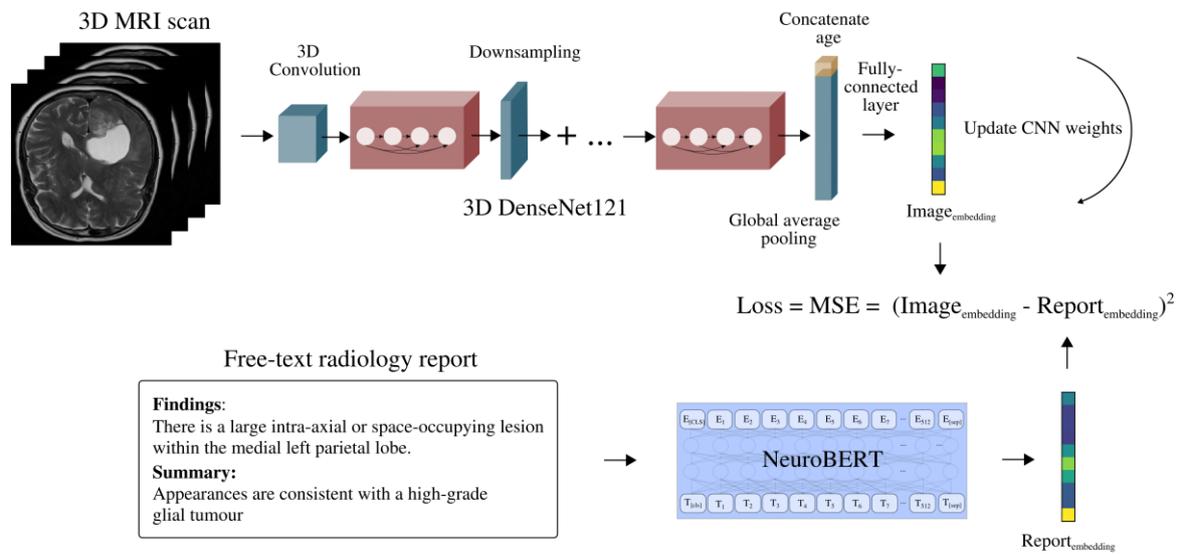

**Figure 4:** *Overview of our computer vision training procedure. Each single-sequence image encoder, based on the DenseNet121 CNN architecture, was trained to map 3D brain MRI scans to the embeddings generated by NeuroBERT from the associated neuroradiology reports. This was achieved by minimising the mean square error (MSE) loss between the report and image embeddings.*

### 3.1.3 Brain abnormality detection in unreported scans

At test-time, our text-vision framework can be used to detect abnormalities in unreported brain MRI examinations by treating the problem as a text-image similarity task and scoring scans against suitable query sentences (Fig. 1). To do this, we first feed a test image into our image encoder to generate an image embedding. Next, one or more (depending on the application) expert-derived query sentences are fed into NeuroBERT to generate query text embeddings. A similarity score between 0 and 1 is then obtained by calculating the cosine similarity between the image embedding and each text embedding, and clipping negative values (i.e., setting these equal to 0). These scores are then used in the same way as traditional probabilities (e.g., to make classification decisions based on thresholds or generate receiver operating characteristic (ROC) curves etc.).

### 3.1.4 Model explainability

In this section, we introduce a simple method to scrutinise our framework's predictions based on guided backpropagation (Springenberg et al., 2014). Initially, we compute the derivative of the similarity score for a given query sentence, and backpropagate this through the image encoder, retaining only the positive error signals (hence 'guided'). The result is a gradient array highlighting the regions which, if changed slightly, would substantially alter the text-image similarity score.

By default, guided backpropagation generates heatmaps matching the dimensionality of the input images. However, scrolling through a stack of slices looking for 'hot spots' might be too time-consuming for radiologists in the clinic. Therefore, to increase utility we additionally generate slice-wise saliency lineouts by computing the maximum gradient per slice. We then automatically display the heatmap for the most influential slice by taking the argmax of this 1D function. Following Smilkov et al. (2017), we supress spuriously high gradients which can occur due to image noise by repeating the guided backpropagation procedure 50 times, each time adding Gaussian noise (mean = 0, standard deviation = 0.1) to the input image, and then utilise the mean of this aggregated gradient array.

### 3.2 Data

All 81,936 consecutive MRI head examinations for patients aged 18 years and above performed in the UK at King's College Hospital National Health Service (NHS) Foundation Trust (KCH) and Guy's and St Thomas' NHS Foundation Trust (GSTT) between 2008-2019 were obtained for this study. KCH and GSTT are two large NHS hospitals covering all neurological, neurosurgical and multi-system medical conditions, where patients are from a wide geographical catchment, and are broadly representative of different social strata and ethnicity with up to 40% non-white. The MRI scans were performed on Signa 1.5 T HDX (General Electric Healthcare, Chicago, US), Aera 1.5 T (Siemens, Erlangen, Germany), Avanto 1.5 T (Siemens), Ingenia 1.5 T (Philips Healthcare, Eindhoven, Netherlands), Intera 1.5 T (Philips) or Skyra 3 T (Siemens) scanners. The text of the corresponding 81,936 routine neuroradiology reports produced by expert neuroradiologists (UK consultant grade; US attending equivalent) were extracted from the Computerised Radiology Information System (CRIS) (Healthcare Software Systems, Mansfield, UK). These reports were largely unstructured and typically comprised 5-10 sentences of image interpretation, sometimes with comments regarding the patient's clinical history, and recommended actions for the referring doctor. All data were de-identified. The UK National Health Research Authority and Research Ethics Committee approved this retrospective study.

To maximise clinical utility, we sought to optimise our framework for use with the most commonly performed MRI sequences. At KCH and GSTT, these were axial $T_2$-weighted, axial diffusion-weighted imaging (DWI), coronal $T_2$-FLAIR, axial gradient recalled echo (GRE) $T_2^*$-weighted, coronal $T_1$-weighted, axial $T_1$-weighted post-contrast, axial $T_2$-FLAIR, and coronal $T_1$-weighted post-contrast images (Fig. B1 in Appendix B). Therefore, only those examinations which included at least one of these sequences were included for model training and testing.

### 3.2.3 Training and testing datasets

Prior to model training, several test sets with 'reference-standard' labels were generated to evaluate the classification performance of our framework. All labelling was performed by three expert neuroradiologists (UK consultant grade; US attending equivalent) who assessed all available image sequences, with a consensus classification decision made in cases where there was any disagreement.

First, a test set of 800 examinations was labelled as either radiologically 'normal' or 'abnormal', following comprehensive pre-defined criteria (Appendix C) (Wood et al., 2022). Briefly, findings which could generate a downstream clinical intervention were labelled as 'abnormal' (referral for case discussion at a multi-disciplinary team meeting was included as the minimal intervention). The 'abnormal' category included findings which were 'excessive for age' (e.g., excessive volume loss, excessive small vessel disease seen on $T_2$-weighted images defined using the Fazekas scale etc.). Next, five 'specialised' reference-standard test sets of 200 examinations each were generated: 'acute stroke' versus 'no acute stroke', 'multiple sclerosis' versus 'no multiple sclerosis', intracranial haemorrhage' versus 'no intracranial haemorrhage', 'meningioma' versus 'no meningioma', and 'hydrocephalus' versus 'no hydrocephalus'. Each dataset consisted of 100 examinations containing the abnormality of interest (the 'positive' category), and 100 examinations where this abnormality was absent (the 'negative' category). Importantly, the negative category (i.e., 'no acute stroke', 'no multiple sclerosis' etc.) always contained a selection of normal examinations as well as a random selection of any abnormal examinations, reflecting real-world clinical practise. These five categories were specifically chosen for their clinical significance and the diversity of neuroradiological presentations they represent, ensuring a robust and relevant evaluation of our framework's image classification performance.

In addition to these classification testing sets with reference standard labels, a large hold-out set of 5000 unlabelled examinations was set aside to evaluate our model's ability to perform visual-semantic image retrieval (see section 3.3). All remaining examinations were then split into unlabelled training (85%) and validation (15%) datasets to facilitate text-vision model development. These data, as well as the reference-standard test data, were split at the level of patients to avoid 'data leakage' (i.e., all examinations of the same patient, including follow-up imaging, were only present in a single dataset). Further dataset information is provided in Table 1.

**Table 1:** *Training, validation, and testing datasets used for the different experiments in this study.*

| MRI sequence | N train | N valid | N test | | | | | | |
|---|---|---|---|---|---|---|---|---|---|
| | | | Normal/ abnormal | Acute stroke | Multiple sclerosis | Haemorrhage | Meningioma | Hydrocephalus | Database search |
| Axial $T_2$ | 50,523 | 6,315 | 800 | 200 | 200 | 200 | 200 | 200 | 5000 |
| Axial DWI | 42,220 | 5,246 | 718 | 200 | 178 | 183 | 188 | 183 | 4398 |
| Coronal $T_2$-FLAIR | 28,351 | 3,518 | 451 | 121 | 118 | 132 | 138 | 122 | 2811 |
| Axial GRE | 20,613 | 2,605 | 384 | 116 | 102 | 200 | 100 | 110 | 2102 |
| Coronal $T_1$ | 11,080 | 1,365 | 169 | 78 | 81 | 73 | 71 | 92 | 1190 |
| Axial $T_1$ (post-contrast) | 10,695 | 1,329 | 151 | 74 | 112 | 70 | 78 | 61 | 1015 |
| Axial $T_2$-FLAIR | 9,582 | 1,236 | 136 | 59 | 55 | 63 | 61 | 58 | 934 |
| Coronal $T_1$ (post-contrast) | 7,736 | 976 | 109 | 48 | 88 | 39 | 32 | 29 | 878 |

### 3.2.1 Brain MRI pre-processing

Minimal image pre-processing was performed. 3D brain MRI scans of arbitrary resolution and dimensions, stored as Digital Imaging and Communications in Medicine (DICOM) files, were converted into NIfTI format, resampled to common voxel sizes and dimensions (1 mm$^3$), and then cropped or padded to a common array size (180 mm x 180 mm x 180 mm). Each image's intensity was normalised by subtracting the mean and dividing by the standard deviation.

Computationally expensive pre-processing steps that could potentially limit real-time clinical utility, such as bias-field correction and any form of spatial registration (including the realignment of angled scans), were avoided. Skull-stripping was not performed to allow for the detection of important extra-cranial abnormalities.

All pre-processing was performed using open-access Python-based libraries. DICOM files were loaded using pydicom, and then converted to NIfTI format using dcm2niix (Li et al.,

2016). NiBabel and numpy (Harris et al., 2020) were used to load and manipulate NIfTI files, and Project MONAI (Cardoso et al., 2022) was used to resample, resize, and normalise each image.

Scripts to enable readers to reproduce these pre-processing steps are available at https://github.com/MIDIconsortium/Neuroimage_preprocessing.

### 3.3 Experiments

### 3.3.1 NeuroBERT training

All language modelling tasks were carried out using Huggingface 4.12.5 (Wolf et al., 2020) with two NVIDIA RTX 2080 graphics processing units (GPUs). We performed WordPiece Tokenizer training, MLM, and RSM using all neuroradiology reports in the unlabelled training dataset (N = 50,523). The vocabulary size was set to 10,000 for tokenizer training, with a minimum token frequency of 10 across our unlabelled training corpus. For MLM, we used a sequence 'block' length of 128, and randomly masked 15% of tokens. The model was trained for 250 epochs using mini-batches of 64 sequence blocks. Early stopping was employed, monitoring the loss on the unlabelled validation set (N = 6315). The model configuration with the lowest validation loss served as the initial configuration for radiology section matching training.

In the case of RSM, we used a rule-based regular expression to separate Findings and Summary sections from free-text neuroradiology reports, and randomly sampled batches of true and false Findings/Summary pairs (examples of true and false Findings/Summary pairs appear in Appendix D). We set the mini-batch size to 16, and the learning rate to $10^{-6}$, incorporating a linear warm-up ramp of 100 steps. The training objective was the MSE between the cosine similarity score and the ground-truth label (1 for true pairs, 0 for false pairs). After training for 50 epochs, the model configuration yielding the lowest validation set loss was adopted as the final NeuroBERT weights.

### 3.3.2 CNN training

The DenseNet121 models used in this study were adapted from the Project MONAI implementation. All vision modelling experiments were conducted using PyTorch 1.7.1 (Paszke et al., 2019) with the two NVIDIA GPUs. The Adam optimizer (Kingma and Ba, 2014) was used with an initial learning rate of $10^{-4}$ which was reduced by a factor of 10 after every 5 epochs without validation loss improvement.

### 3.3.3 Classification performance evaluation

Our framework's classification performance was evaluated in two ways: i) through a 'normal' versus 'abnormal' classification task relevant for triage applications; and ii) through five specialised diagnostic classification tasks, specifically 'acute stroke' versus 'no acute stroke', 'multiple sclerosis' versus 'no multiple sclerosis', 'intracranial haemorrhage' versus 'no intracranial haemorrhage', 'meningioma' versus 'no meningioma', and 'hydrocephalus' versus 'no hydrocephalus'. In both cases, we refer to this as *zero-shot classification*, given that our text-vision model has not been explicitly trained on these categories, but instead leverages the inherent contextual understanding gained from self-supervised learning on a diverse set of neuroradiology reports and imaging data. In the 'normal' versus 'abnormal' classification, all available sequences for each examination in our reference-standard test set were scored against the sentence 'there are normal intracranial appearances'. We then subtracted the resulting text-image 'normality scores' from 1 to generate an 'abnormality score' in the range [0,1] for each sequence. We reported the classification performance for each MRI sequence individually and investigated an ensemble strategy which returns the highest single-sequence abnormality score for each examination, mirroring how a radiologist might approach abnormality detection.

For the specialized classification tasks (acute stroke, multiple sclerosis, intracranial haemorrhage, meningioma, hydrocephalus) all available sequences for each examination in our five specialized reference-standard test sets were scored against the expert-derived sentences 'there is restricted diffusion in keeping with acute infarction', 'appearances meet the McDonald criteria for multiple sclerosis', 'appearances are in keeping with a recent bleed', 'appearances are in keeping with a meningioma', and 'appearances are in keeping with hydrocephalus', respectively. Here, we utilized the ensemble approach described above to generate examination-level predictions (i.e., we took the highest single-sequence text-image similarity score as the final score for each examination).

The performance of the framework in all cases was quantified using the area under the ROC curve (AUC), along with the macro averages (i.e., the average across the positive and negative classes) of precision, recall, and F1-score. DeLong's test (DeLong et al., 1988) was used to test the statistical significance of differences in AUC scores between the single sequence and ensemble approach, and saliency lineouts and heatmaps were generated using guided backpropagation to visualise influential image regions.

### 3.3.4 Visual-semantic image retrieval

Visual-semantic image retrieval is a technique that involves identifying and retrieving images from a dataset that closely align with given textual descriptors. This approach finds potential applications in clinical decision support, for example, by showcasing historical instances of pertinent findings, and in radiologist training, by providing an interactive, image-centric study resource for specific medical conditions.

To evaluate our framework's capacity for visual-semantic image retrieval, we employed our large, unlabelled hold-out test set of 5000 examinations. Three expert neuroradiologists selected 7 radiological findings relevant to neuroradiology reporting and covering a wide range of scenarios (glioma, Alzheimer's disease, pineal cyst, metastases, post-surgical resection cavity, haematoma, and vestibular schwannoma) and formulated corresponding text queries for each finding ('the findings are in keeping with a glioma', 'there is volume loss in keeping with Alzheimer's disease', 'the appearances are consistent with a pineal cyst', 'the appearances are in keeping with metastatic disease', 'there is a post-surgical resection cavity', 'there is a haematoma', and 'the findings are consistent with a vestibular schwannoma'). Each query was then encoded using NeuroBERT, and the images in the large, hold-out database search test set (N = 5000) were encoded using the relevant single-sequence CNN encoders. All candidate images were ranked based on their cosine similarity to the query, in decreasing order. The top 15 images for each query were selected. We reported our findings using the precision@K metric (where K = 15), a common measure of retrieval effectiveness, calculated by dividing the number of relevant images retrieved by the total number of images retrieved (in this case 15).

## 4. Results

### 4.1 'Normal' versus 'abnormal' classification

Accurate classification of scans as either 'normal' or 'abnormal' was observed for all single-sequence models (Table 2), with best performance achieved using axial $T_2$-weighted scans (AUC = 0.925). Examples of abnormalities correctly identified by our different models, along with the accompanying saliency lineouts and heatmaps, are shown in Figure 5.

**Table 2:** *Zero-shot classification performance (AUC, macro-average precision, macro-average recall, and macro-average F1-score) of our single-sequence vision models on the 'normal' versus 'abnormal' binary classification task. Accurate classification (AUC > 0.75) was achieved for all single-sequence models, although an ensemble model which returned the highest available single-sequence score for each examination outperformed the best single-sequence model. Best performance shown in bold.*

| MRI sequence | AUC | Precision | Recall | F1 score |
|---|---|---|---|---|
| Axial $T_2$ | 0.925 | 0.882 | 0.858 | 0.866 |
| Coronal $T_2$-FLAIR | 0.920 | 0.842 | 0.859 | 0.849 |
| Axial DWI | 0.902 | 0.832 | 0.869 | 0.839 |
| Axial $T_2$-FLAIR | 0.886 | 0.798 | 0.849 | 0.817 |
| Axial $T_1$ (post-contrast) | 0.885 | 0.775 | 0.800 | 0.787 |
| Axial GRE | 0.855 | 0.825 | 0.828 | 0.820 |
| Coronal $T_1$ | 0.847 | 0.727 | 0.818 | 0.761 |
| Coronal $T_1$ (post-contrast) | 0.769 | 0.668 | 0.755 | 0.699 |
| Ensemble | **0.950** | **0.883** | **0.881** | **0.882** |

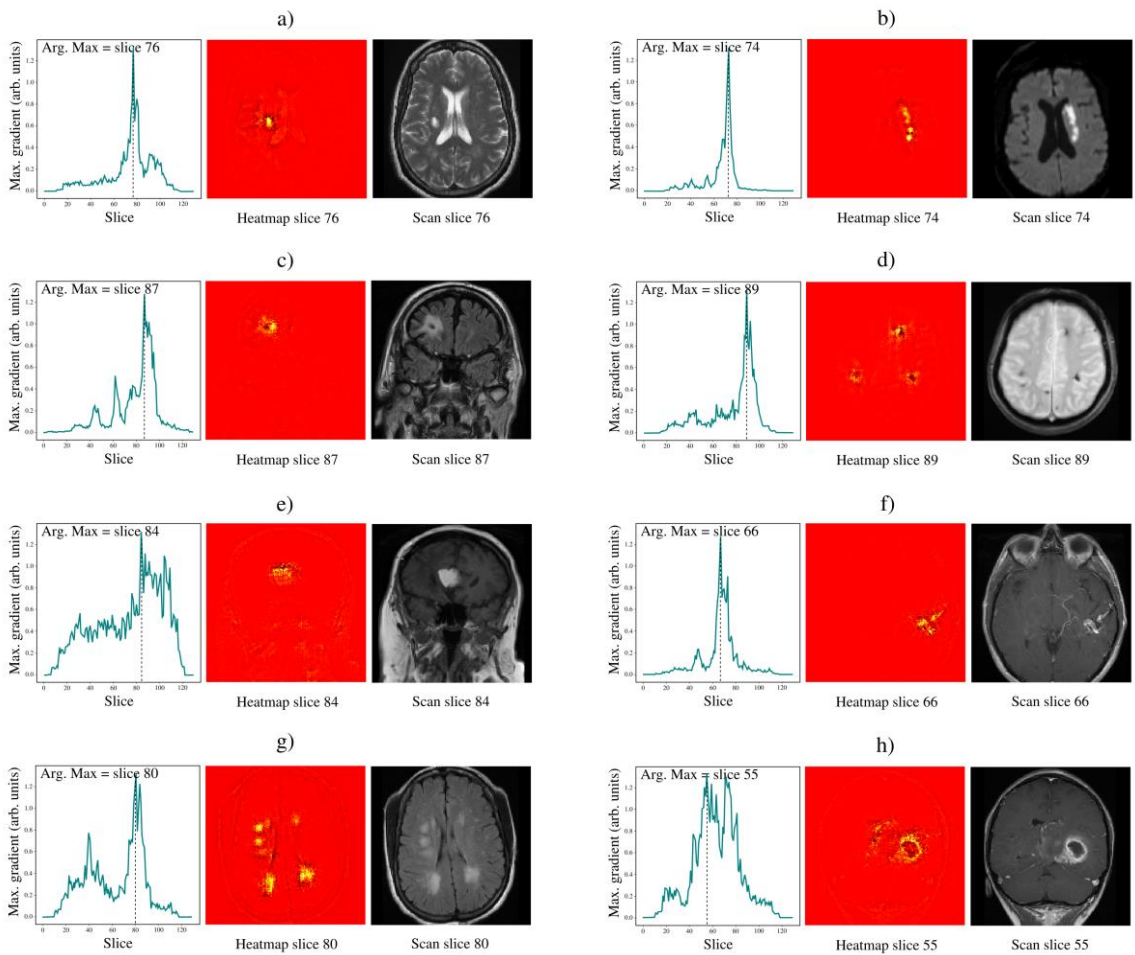

**Figure 5:** *Smooth guided backpropagation enables 'slice-wise' and 'voxel-wise' visualisations of image regions which are most influential for our model's predictions. When applied to scans from the 'normal' versus 'abnormal' test set, accurate localisation of a range of morphologically distinct abnormalities, both across and within slices, is observed for all single-sequence models. Clockwise from top left: a) axial $T_2$-weighted image showing a right corona radiata chronic infarct; b) axial DWI image showing acute infarction in the left caudate and corona radiata; c) coronal $T_2$-FLAIR image showing focal encephalomalacia in the right frontal lobe due to previous infarction; d) axial GRE image showing numerous cerebral microhaemorrhages; e) coronal $T_1$-weighted image showing a right frontal lipoma; f) axial $T_1$-weighted (post-contrast) image showing a left posterior temporal arteriovenous malformation; g) axial $T_2$-FLAIR image showing severe small vessel disease; h) coronal $T_1$-weighted (post-contrast) image showing a high-grade glioma. In each example, the left panel shows the saliency across different slices; the middle panel shows the heatmap for the most influential slice; and the right panel shows the corresponding scan slice. Note: using radiological left and right.*

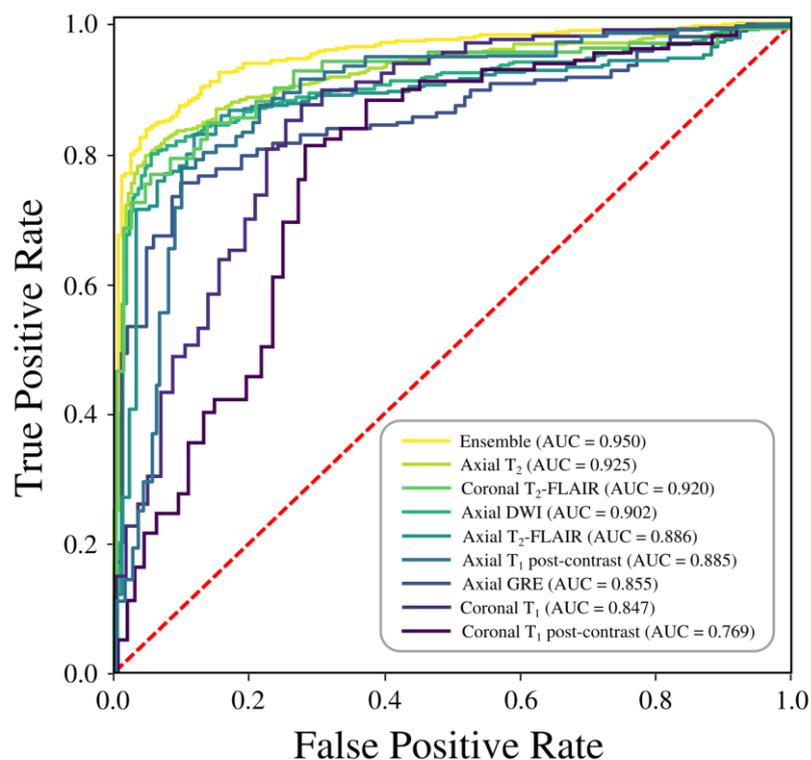

**Figure 6:** *Receiver operating characteristic curves for our 8 single-sequence text-vision models on the 'normal' versus 'abnormal' binary classification task, along with an ensemble approach which returns the highest available single-sequence abnormality score for each examination.*

'Normal' versus 'abnormal' classification accuracy was improved when using an ensemble strategy which returned the highest single-sequence abnormality score for each examination ($p < 0.05$) (Fig. 6). This approach ensured that examinations containing abnormalities which are conspicuous on certain sequences and less visible on others were correctly classified (Fig. 7).

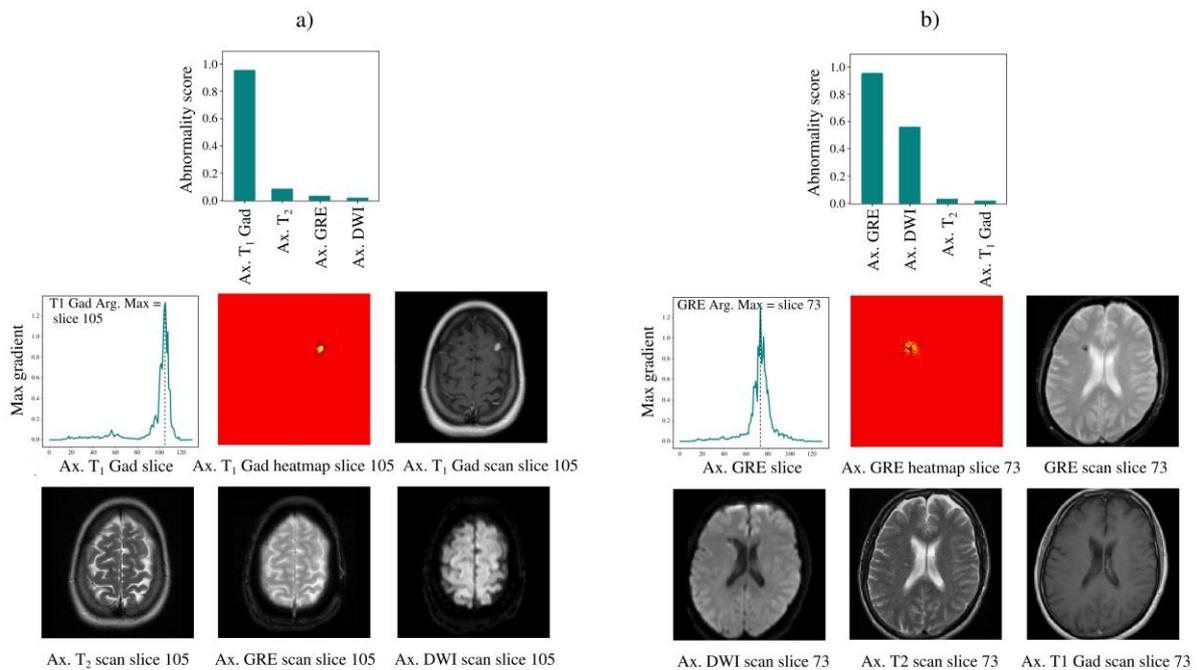

**Figure 7:** *Visualisation of our ensemble approach to brain abnormality detection. Abnormality scores are calculated for all available MRI sequences using the relevant single-sequence model. These scores are then ranked from high to low and displayed as a bar chart (top row), with the highest score taken as the examination-level prediction. Saliency lineouts and heatmaps highlighting the most influential image regions for the most important sequence are also generated (middle row). Our ensemble approach ensures that examinations containing abnormalities which are conspicuous on certain sequences and less visible on others are correctly classified. For example, the examination in panel a) contains a left-sided metastasis which is visible only on the contrast-enhanced $T_1$-weighted sequence. Likewise, the examination in panel b) contains a small right-sided microhaemorrhage which is visible only on the GRE and DWI sequences. This differential visibility is reflected in the sequence abnormality scores. Note: using radiological left and right.*

### 4.2 Specialised abnormality detection

In addition to 'normal' versus 'abnormal' categorisation, our text-vision framework is able to classify examinations into more specialised classes. Accurate classification was observed for all five categories considered: 'acute stroke' versus 'no acute stroke' (AUC = 0.943), 'multiple sclerosis' versus 'no multiple sclerosis' (AUC = 0.935), 'intracranial haemorrhage' versus 'no intracranial haemorrhage' (AUC = 0.890), 'meningioma' versus 'no meningioma' (AUC = 0.882), and 'hydrocephalus' versus 'no hydrocephalus' (AUC = 0.810) (Fig. 8, Table 3). Figure 9 presents examples of correctly classified examinations for these five categories, along with the corresponding saliency lineouts and heatmaps.

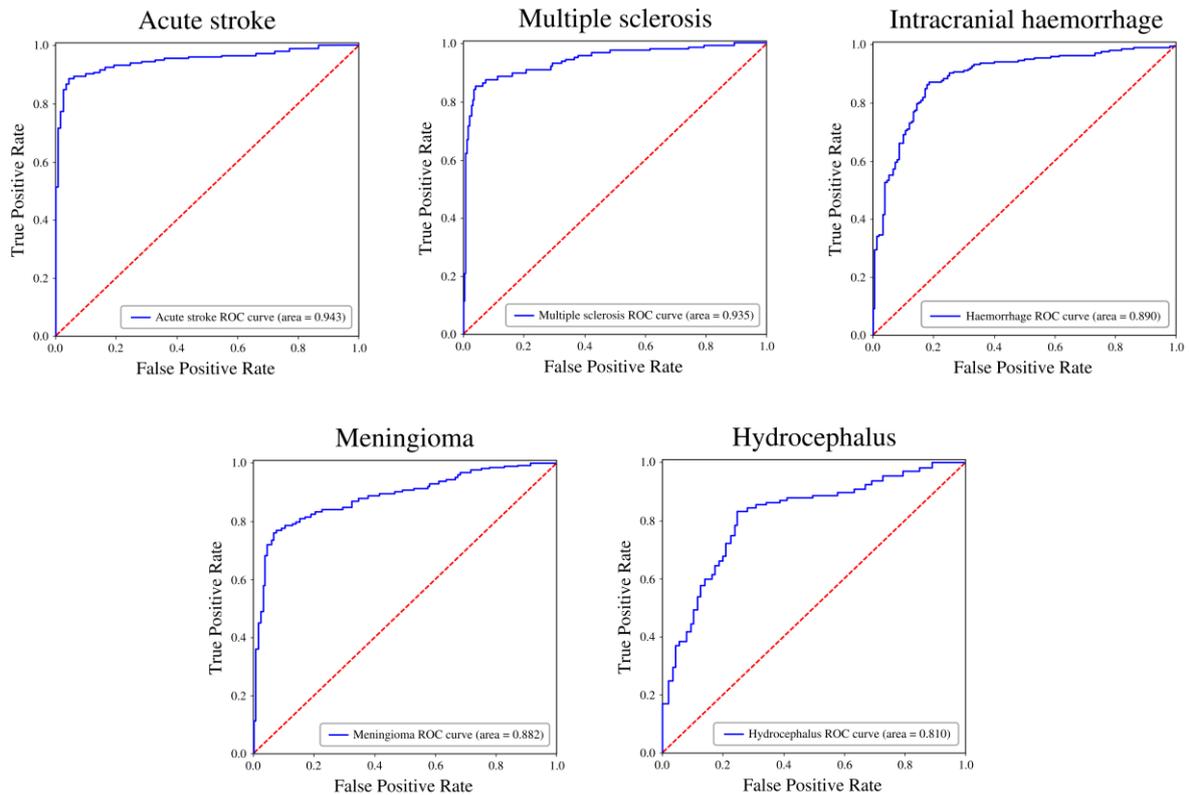

**Figure 8:** *Receiver operating characteristic curve when utilising our ensemble approach to classify brain MRI examinations as either 'acute stroke' or 'no acute stroke' (top left), 'multiple sclerosis' or 'no multiple sclerosis' (top middle), 'intracranial haemorrhage' or 'no intracranial haemorrhage' (top right), 'meningioma' or 'no meningioma' (bottom left), and 'hydrocephalus' or 'no hydrocephalus' (bottom right).*

**Table 3:** *Zero-shot classification performance (AUC, macro-average precision, macro-average recall, and macro-average F1-score) for the five specialised abnormality classification tasks. Accurate classification (AUC $\geq$ 0.81) was achieved for all pathologies.*

| Zero-shot classification task | AUC | Precision | Recall | F1 score |
|---|---|---|---|---|
| 'Acute stroke' versus 'no acute stroke' | 0.943 | 0.944 | 0.939 | 0.941 |
| 'Multiple sclerosis' versus 'no multiple sclerosis' | 0.935 | 0.916 | 0.904 | 0.910 |
| 'Intracranial haemorrhage' versus 'no intracranial haemorrhage' | 0.890 | 0.845 | 0.836 | 0.841 |
| 'Meningioma' versus 'no meningioma' | 0.882 | 0.841 | 0.835 | 0.835 |
| 'Hydrocephalus' versus 'no hydrocephalus' | 0.810 | 0.781 | 0.786 | 0.779 |

a)

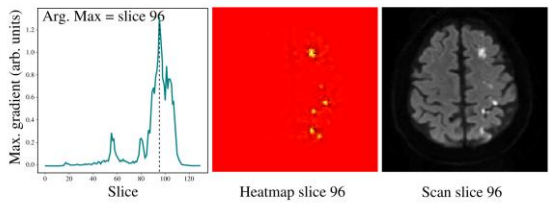 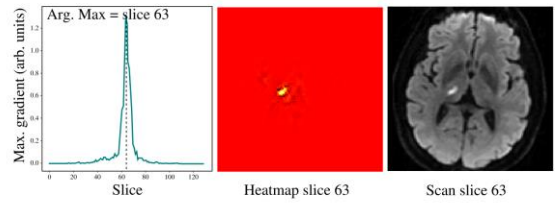

b)

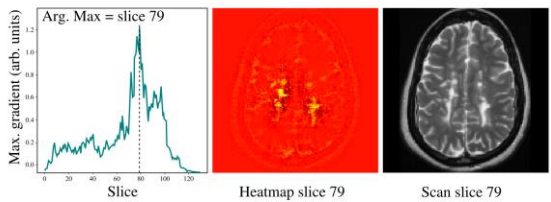 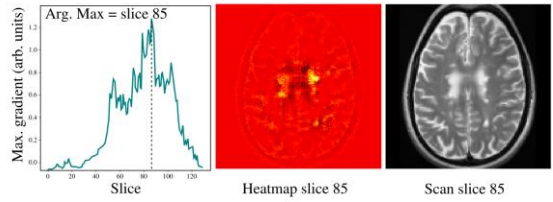

c)

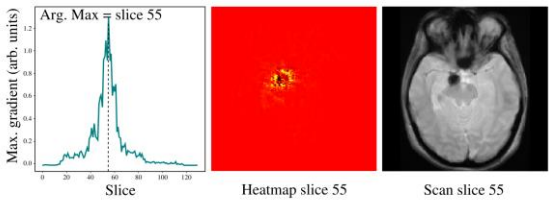 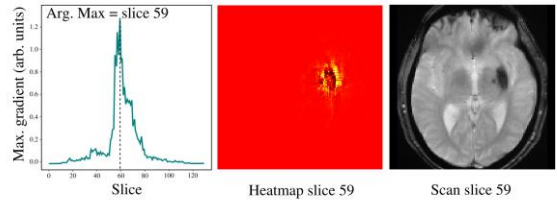

d)

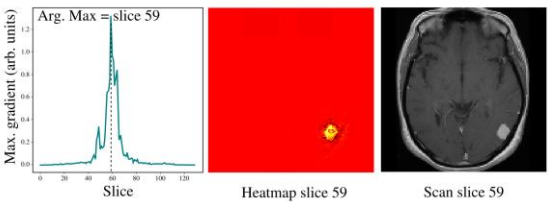 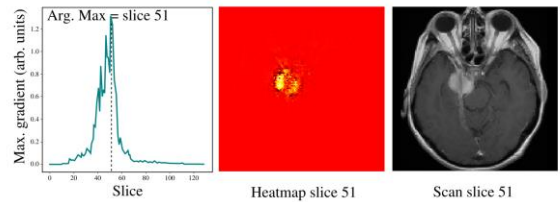

e)

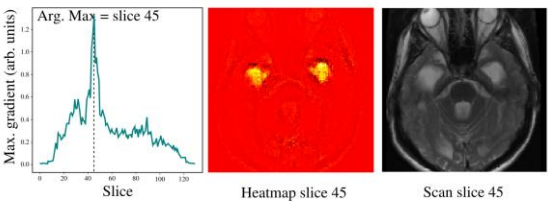 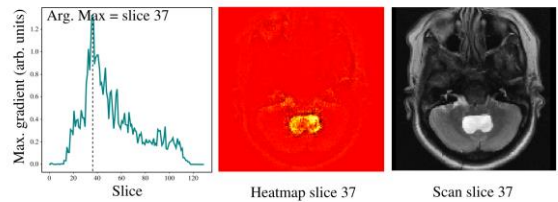

**Figure 9:** *Smooth guided backpropagation applied to scans from the five specialised classification test sets. a) acute stroke (acute infarction on DWI imaging); b) multiple sclerosis (periventricular Dawson's fingers on $T_2$-weighted imaging); c) intracranial haemorrhage (susceptibility from blood products on GRE imaging); d) meningioma (avid and homogenous extra-axial enhancement on $T_1$-weighted (post-contrast) imaging); e) hydrocephalus (enlarged ventricles on $T_2$-weighted imaging).*

### 4.3 Visual-semantic database search

Accurate image retrieval was achieved for all seven pathologies considered (average precision@15 = 0.842) (Table 4). Examples of correctly retrieved images for each category are shown in Fig. 10, along with the corresponding neuroradiology report (which was not used for retrieval); Fig. E1 in appendix E shows examples of the few cases where our text-vision model retrieved images which do not contain the pathology requested; in each case, ambiguous or morphologically similar pathologies were returned.

**Table 4:** *Visual-semantic database search results. Accurate retrieval of images from textual description prompts was observed for all seven abnormality categories examined (average precision@15 = 0.840).*

| Abnormality | Precision@15 |
|---|---|
| Post-surgical resection cavity | 0.933 |
| Haematoma | 0.917 |
| Pineal cyst | 0.867 |
| Glioma | 0.846 |
| Vestibular schwannoma | 0.833 |
| Metastasis | 0.764 |
| Alzheimer's disease | 0.722 |

a)

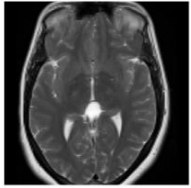 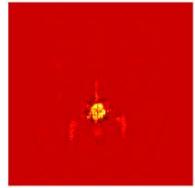

**Findings:**
There is a simple pineal cyst measuring 17 mm in diameter. No convex features. No solid components and no restricted diffusion. There is mild anterior displacement of the tectal plate. There is no abnormal enhancement seen.

**Summary:**
Simple 17 mm pineal cyst without complex features. The ventricles are non-dilated and there is no evidence of hydrocephalus.

b)

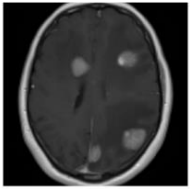 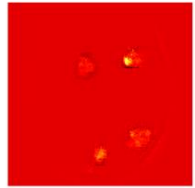

**Findings:**
There are numerous large enhancing masses noted. Some of the lesions are intra-axial including a 2.7 cm left frontal mass. There is a 2.8 cm right periventricular lesion centred in the head of the right caudate with some suggestion of adjacent subependymal enhancement along the margin of the right frontal horn. Several other lesions appear dural based including a left 3.2 centimetre parietal mass and left medial parieto-occipital parietal mass, abutting the postero aspect of the superior sagittal sinus, but without clearly encroaching on the flow-void.
Further smaller lesions are noted in the left temporal lobe and right frontal lobe.

**Summary:**
There are numerous large intra and extra-axial masses involving the cerebral and cerebellar hemispheres, which in the clinical context is likely to represent metastatic deposits.

c)

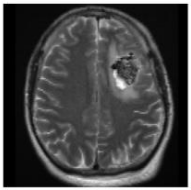 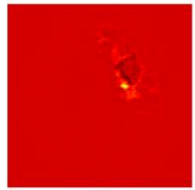

**Findings:**
There is a left fronto-parietal mini craniotomy. The left frontal lobe peripheral resection cavity is noted with associated T2 hyperintense/T1 isointense fluid and multiple foci of low signal consistent with air and gliadel wafers. There is T2 hyperintensity within the white matter of the left frontal lobe adjacent to the resection. There is some minimal linear areas of T1 hyperintensity at the deep and posterior margin of resection.

**Summary:**
The left frontal glioblastoma resection cavity is noted. Although potentially masked by minor areas of T1 hyperintensity, there is no residual gadolinium enhancement evident at this site of resection. Non specific adjacent T2 hyperintensity is noted within the white matter.

d)

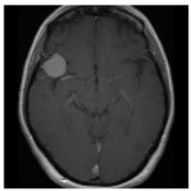 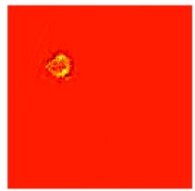

**Findings:**
There are stable appearances to the meningioma within the right Sylvian fissure since the previous study (maximum dimensions = 2.4 cm). There is only minor distortion of the adjacent anterior temporal lobe - brain architecture is otherwise well preserved. No further intracranial abnormality is shown.

**Summary:**
No interval change to the known right-sided meningioma.

e)

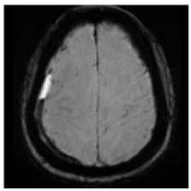 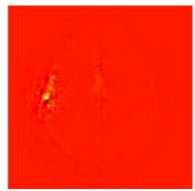

**Findings:**
Comparison is made with the CT head examination dated [ANONYMISED]. There remains a mixed-signal shallow right cerebral convexity subdural collection. There is smooth enhancement of the dura. No focal mass lesion, abnormal signal or focal enhancement is demonstrated to suggest an underlying cause for the recurrent haemorrhage. Slight mass effect persists (partial sulcal effacement is again noted) but there is no midline shift and the basal cisterns are patent.

**Summary:**
Right cerebral convexity subdural collection, as described.

f)

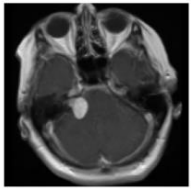 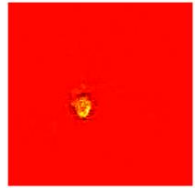

**Findings:**
There is a right cerebello pontine angle cistern and IAM lesion which is of intermediate T1 and T2 signal, undergoing moderate and slightly heterogeneous gadolinium enhancement. It measures 2.1cm parallel to the petrous pyramid x 1.3cm (perpendicular to the petrous pyramid). In its extra meatal portion it measures 2.2cm (cranio caudal). It does not quite reach the fundus of the right IAM. There is moulding of the right middle cerebellar peduncle. There is some patchy hemispheric white matter T2 hyperintensity. The right cerebello pontine angle cistern lesion distorts the root entry zone of the right trigeminal nerve. There is no other lesion in the line of the right trigeminal nerve.

**Summary:**
There is a right cerebello pontine angle cistern and IAM lesion consistent with a vestibular schwannoma. There is associated impingement on the right trigeminal nerve.

g)

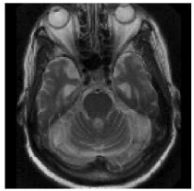 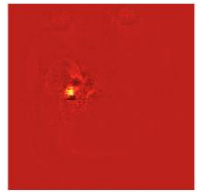

**Findings:**
There is bilateral medial temporal volume loss-particularly marked on the right. There is a background of age-related involutional change and mild to moderate small vessel ischaemic change.

**Summary:**
Bilateral medial-temporal volume loss, consistent with Alzheimer's type neuro-degeneration.

**Figure 10:** *Examples of correctly retrieved images using our visual-semantic database search technique. Heatmaps generated by guided backpropagation are also shown, along with the corresponding neuroradiology report for each examination (which was not used for image retrieval). a) retrieval task: find image examples of a pineal cyst; correctly retrieved example: a 17mm pineal cyst; b) retrieval task: find image examples of metastatic deposits; correctly retrieved example: numerous cerebral metastatic deposits; c) retrieval task: find image examples of a post-surgical resection cavity; correctly retrieved example: a left frontal post-surgical resection cavity; d) retrieval task: find image examples of a meningioma; correctly retrieved example: a right sylvian fissure meningioma; e) retrieval task: find image examples of a haematoma; correctly retrieved example: a right convexity subdural haematoma; f) retrieval task: find image examples of a vestibular schwannoma; correctly retrieved example: a right cerebello-pontine vestibular schwannoma; g) retrieval task: find image examples of Alzheimer's disease; correctly retrieved example: marked right-sided hippocampal volume loss in keeping with Alzheimer's disease. Note: using radiological left and right.*

**Discussion**

The growing demand for brain MRI examinations and a global shortfall of radiologists are exerting substantial pressure on healthcare systems (NHS, 2021)(Wood et al., 2021). This surge in workload underscores the need for AI-driven triage and decision-support systems to help reduce the burden on radiologists. However, the development of computer vision modes is impeded by a critical bottleneck: growing clinical demands make it impractical to dedicate radiologists' time to the manual labelling of large and clinically diverse training datasets (Benger et al., 2023). To address these challenges, we have developed a self-supervised text-vision framework. Our approach uses unlabelled brain MRI scans and the information in their associated neuroradiology reports to learn to identify abnormalities. This method eliminates the need for large, manually annotated datasets.

To ensure the robustness and applicability of our text-vision framework, we conducted a rigorous evaluation across various classification and image retrieval tasks. Using an ensemble strategy, which selects the highest abnormality score from all available single-sequence models, we achieved an AUC of 0.950 on the 'normal' versus 'abnormal' classification task. This significantly outperformed individual sequence predictions which achieved a mean AUC of 0.874 across all single-sequence models. This ensemble approach not only improves classification accuracy, but also closely mirrors the real-world diagnostic process of radiologists, who assess multiple sequences to form a complete understanding of a patient's scan.

In specialized classification tasks tailored for refined triage and decision support, the framework achieved a mean AUC of 0.892 across five distinct pathology categories (acute stroke, multiple sclerosis, intracranial haemorrhage, meningioma, and hydrocephalus). It is important to note that these results were achieved through *zero-shot* classification; in other words, the models were not explicitly trained to perform these tasks. The potential for enhancing classification performance by fine-tuning with smaller, labelled datasets is a promising direction for future research.

In the task of textual query-based image retrieval, the framework achieved a mean precision of 0.840 across 7 clinically relevant pathologies (pineal cyst, glioma, haematoma, vestibular schwannoma, Alzheimer's disease, metastasis, and post-surgical resection cavity). By allowing radiologists to pull up similar historical cases based on textual descriptors, this feature could aid in the interpretation of challenging cases and serve as a valuable educational tool or diagnostic decision support.

Another key strength of our study is the flexibility afforded by zero-shot classification through text-image similarity scoring. Traditional computer vision models are often constrained to a predefined set of categories, necessitating new training examples for each new classification task. This rigidity poses a considerable challenge in neuroradiology, where clinical requirements and diagnostic categories continually evolve. Our text-vision framework overcomes this limitation as it has the capability to generate similarity scores for a wide range of image findings, adaptable to various clinical scenarios. This flexibility enables more dynamic and practical classification, better suiting the varying needs of clinical practice and enhancing the utility for radiologists.

Finally, our framework includes a practical interpretability feature based on guided backpropagation, applied - for the first time to the best of our knowledge - to text-image similarity calculations. This method helps identify the most influential image regions for the model's predictions and, by automatically pinpointing key slices in MRI scans, supports radiologists in quick decision-making while offering an additional layer of quality assurance.

A promising future application of our text-vision framework, not examined in the current study, is its use in detecting discrepancies in provisional radiology reports. Figure F1 in Appendix F illustrates this concept. This method has the potential to substantially enhance the accuracy of initial radiological assessments, reducing the risk of diagnostic errors; however, this application would require extensive clinical validation to ensure its efficacy and safety, a process which was beyond the scope of the current study.

There are some limitations of our study to consider. First, although our ensemble approach to brain abnormality detection involves integrating predictions from multiple MRI sequences, it ultimately treats each sequence independently. This approach will be adequate for detecting and differentiating most abnormalities; however, whilst we acknowledge it is a very rare scenario, accurate diagnosis occasionally hinges on the presence of combined findings across multiple sequences. An example of this is brain abscesses which, whilst rare, is a critical and often urgent medical condition. Abscesses typically show ring enhancement on $T_1$ post-contrast imaging, a feature that can also be seen in high-grade tumours, making differentiation based solely on this criterion sometimes challenging. However, the combination of $T_1$ post-contrast enhancement and restricted diffusion on DWI images can be key in accurately distinguishing abscesses from tumours; as such, it is possible that our independent-sequence approach may encounter difficulties in differentiating between such abnormalities.

A further, related, limitation is that our single-sequence computer vision models are trained to map individual scans to examination-level embeddings. Potentially, this could be problematic when a report describes findings which are conspicuous on one MRI sequence, but subtle or even invisible on another (e.g., small vessel disease, which is easily visible on $T_2$ and $T_2$-FLAIR imaging, but is sometimes less discernible on $T_1$-weighted imaging). Ideally, individual computer vision models should be trained to map scans to sequence-specific embeddings which capture the relevant sequence-specific findings in a given report. Nonetheless, we have shown that our ensemble approach can detect abnormalities which are conspicuous on certain sequences and less visible on others, mitigating this issue to a large extent.

Finally, the dependency of our radiology report section matching pre-training task on the quality of radiology reports poses a potential limitation. Inaccuracies or non-informative Findings and Summary sections can impede the model's ability to distinguish true pairs from false pairs effectively, leading to less accurate report embeddings. This, in turn, can negatively impact the computer-vision model performance, as it is trained to map images to these embeddings. A notable challenge arises with follow-up reports, where radiologists may not reiterate all findings, instead opting for phrases like 'stable appearances' or 'no interval change.' Such reports can lack the detailed descriptions necessary for our model to learn robust embeddings. For follow-up reports, comparison with earlier reports and images might sometimes be necessary, which likely requires an alternative framework trained to detect subtle changes over time.

In conclusion, we have presented a self-supervised text-vision framework which learns to detect clinically relevant abnormalities from unlabelled hospital brain MRI scans, eliminating the need for large, expert-labelled training datasets. Using this framework, we have demonstrated accurate zero-shot classification of brain MRI scans into both general as well as more specialised categories, opening a range of applications including automated triage. In addition, our framework demonstrates potential for clinical decision support and education by retrieving and displaying examples of pathologies from historical examinations that could be relevant to the current case based on textual descriptors.

Din, M., Agarwal, S., Grzeda, M., Wood, D. A., Modat, M., & Booth, T. C. (2023). Detection of cerebral aneurysms using artificial intelligence: a systematic review and meta-analysis. *Journal of NeuroInterventional Surgery*, *15*(3), 262-271.

Din, M., Agarwal, S., Grzeda, M., Wood, D. A., Modat, M., & Booth, T. C. (2023). Detection of cerebral aneurysms using artificial intelligence: a systematic review and meta-analysis. *Journal of NeuroInterventional Surgery*, *15*(3), 262-271.

Fazekas, F., Chawluk, J. B., Alavi, A., Hurtig, H. I., & Zimmerman, R. A. (1987). MR signal abnormalities at 1.5 T in Alzheimer's dementia and normal aging. *American Journal of Neuroradiology*, *8*(3), 421-426.

Frome, A., Corrado, G. S., Shlens, J., Bengio, S., Dean, J., Ranzato, M. A., & Mikolov, T. (2013). Devise: A deep visual-semantic embedding model. *Advances in neural information processing systems*, *26*.

Gulshan, V., Peng, L., Coram, M., Stumpe, M. C., Wu, D., Narayanaswamy, A., … & Webster, D. R. (2016). Development and validation of a deep learning algorithm for detection of diabetic retinopathy in retinal fundus photographs. *Jama*, *316*(22), 2402-2410.

Harris, C. R., Millman, K. J., Van Der Walt, S. J., Gommers, R., Virtanen, P., Cournapeau, D., … & Oliphant, T. E. (2020). Array programming with NumPy. *Nature*, *585*(7825), 357-362.

Huang, G., Liu, Z., Van Der Maaten, L., & Weinberger, K. Q. (2017). Densely connected convolutional networks. In *Proceedings of the IEEE conference on computer vision and pattern recognition* (pp. 4700-4708).

Huang, S. C., Shen, L., Lungren, M. P., & Yeung, S. (2021). Gloria: A multimodal global-local representation learning framework for label-efficient medical image recognition. In *Proceedings of the IEEE/CVF International Conference on Computer Vision* (pp. 3942-3951).

Irvin, J., Rajpurkar, P., Ko, M., Yu, Y., Ciurea-Ilcus, S., Chute, C., … & Ng, A. Y. (2019, July). Chexpert: A large chest radiograph dataset with uncertainty labels and expert comparison. In *Proceedings of the AAAI conference on artificial intelligence* (Vol. 33, No. 01, pp. 590-597).

Johnson, A. E., Pollard, T. J., Berkowitz, S. J., Greenbaum, N. R., Lungren, M. P., Deng, C. Y., … & Horng, S. (2019). MIMIC-CXR, a de-identified publicly available database of chest radiographs with free-text reports. *Scientific data*, *6*(1), 317.

**Appendix A**

**Table A1:** *Comparison of WordPiece tokenization by BERT and NeuroBERT for common neuroradiological terms. By using a dedicated neuroradiological vocabulary, NeuroBERT produces far fewer word breakdowns compared to BERT when applied to domain-specific jargon.*

| Word | BERT | NeuroBERT |
| --- | --- | --- |
| cerebellar | ['ce', 're', 'bell', 'ar'] | ['cerebellar'] |
| intracranial | ['int', 'rac', 'ran', 'ial'] | ['intracranial'] |
| haemorrhage | ['ha', 'em', 'or', 'rh', 'age'] | ['haemorrhage'] |
| aneurysm | ['ane', 'ur', 'ys', 'm'] | ['aneurysm'] |
| pons | ['p', 'ons'] | ['pons'] |
| suprasellar | ['sup', 'rase', 'll', 'ar'] | ['suprasellar'] |
| pituitary | ['pit', 'uit', 'ary'] | ['pituitary'] |
| infarction | ['inf', 'ar', 'ction'] | ['infarction'] |
| hyperintensities | ['hyper', 'int', 'ens', 'ities'] | ['hyperintensities'] |
| parenchyma | ['paren', 'chy', 'ma'] | ['pa', 'renchyma'] |
| trigeminal | ['tr', 'ig', 'em', 'inal'] | ['trigeminal'] |
| hippocampal | ['h', 'ipp', 'ocamp', 'al'] | ['hippocampal'] |
| encephalomalacia | ['ence', 'phal', 'omal', 'acia'] | ['encephalomalacia'] |
| haematoma | ['ha', 'em', 'at', 'oma'] | ['haematoma'] |

| thalamus | ['thal', 'amus'] | ['thalamus'] |
| subarachnoid | ['sub', 'ar', 'ach', 'n', 'oid'] | ['subarachnoid'] |
| cerebral | ['ce', 're', 'bral'] | ['cerebral'] |
| tumour | ['t', 'um', 'our'] | ['tumour'] |
| ischaemic | ['is', 'cha', 'emic'] | ['ischaemic'] |
| meningioma | ['m', 'ening', 'i', 'oma'] | ['meningioma'] |
| periventricular | ['per', 'iv', 'entric', 'ular'] | ['periventricular'] |
| occipital | ['occ', 'ip', 'ital'] | ['occipital'] |

## Appendix B

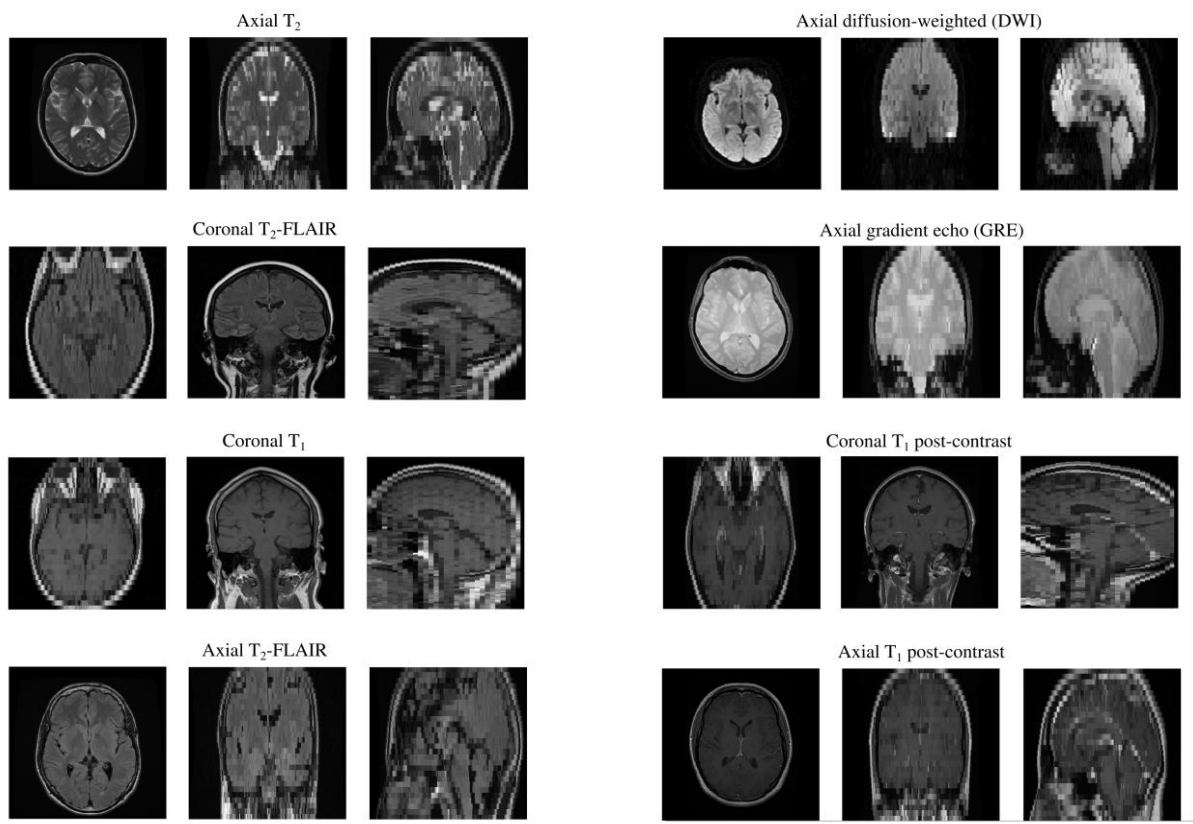

**Figure B1:** *Overview of the brain MRI scans used in this study. Clockwise from top left: axial $T_2$-weighted, axial diffusion-weighted (DWI), axial gradient echo (GRE), coronal $T_1$-weighted (post-contrast), axial $T_1$-weighted (post-contrast), axial $T_2$-FLAIR, coronal $T_1$-weighted, and coronal $T_2$-FLAIR scans.*

**Appendix C.**

Brain abnormality definitions derived by a team of expert neuroradiologists (UK consultant grade; US attending equivalent) during consensus meetings which took place over the course of six months. A more detailed version is available in (Wood et al., 2022a) and (Wood et al., 2022b).

**Binary abnormality definitions**

Abnormal is defined as one or more abnormality described below.

Normal is defined as no abnormality described below.

**Granular (specialised) abnormality definitions:**

**Small vessel disease**

Fazekas and colleagues (Fazekas et al., 1987) gave a classification system for white matter lesions (WMLs) summarised as:

1. Mild - punctate WMLS: Fazekas I

2. Moderate - confluent WMLs: Fazekas II

3. Severe - extensive confluent WMLs: Fazekas III

To create a binary categorical variable from this system, if the report is described as 'unsure', 'normal' or 'mild' this is categorized as normal as this never requires treatment for cardiovascular risk factors. However, if there is a description of moderate or severe WMLs, the report is categorized as abnormal as these cases sometimes require treatment for cardiovascular risk factors.

Included as normal are descriptions of scattered non-specific 'white matter dots' or 'foci of signal abnormality' (unless a more defuse or specific pathology is implied) and small vessel disease described as 'minor', 'minimal' or 'modest'.

Conversely, those cases which are described as 'mild to moderate', 'confluent', or 'beginning to confluence' small vessel disease are treated as abnormal.

Genetic small vessel disease, in particular Cerebral Autosomal Dominant Arteriopathy with Subcortical Infarcts and Leukoencephalopathy (CADASIL), is considered abnormal.

**Mass**

All the following intracranial masses are categorized as abnormal:

− Neoplasms (tumours)

- Intra-axial including all primary and secondary neoplasms
- Extra-axial including all primary and secondary neoplasms
    - Pituitary adenomas included
- Lipomas included

− Tumour debulking or partial resection as this implies residual tumour (note: these are labelled as both 'encephalomalacia' and 'mass' abnormalities)

− Ependymal, subependymal or local meningeal enhancement (non-surgical) in the context of a history of an aggressive infiltrative tumour

− Abscess

− Cysts

- Retrocebellar cyst (mega cisterna magna not included)
- Arachnoid cysts
- Pineal cysts and choroid fissure cysts
- Rathke cleft cysts

– Focal cortical dysplasia, nodular grey matter heterotopia, subependymal nodules and subcortical tubers

– Chronic subdural haematoma or hygroma (i.e., cerebrospinal fluid (CSF) equivalent)

– Perivascular spaces normal unless giant

– MRI examinations for stereotactic surgical planning alone may have very brief reports. In these scenarios it is typically evident from the clinical information provided that there is a mass e.g., surgical planning for glioblastoma.

Note that findings that typically may have minimal clinical relevance when confirmed by a neuroradiology expert, are included in this category e.g., arachnoid cyst. The rationale is that such a finding might generate a referral to a multidisciplinary team meeting for clarification clinical relevance. We consider that a referral to a multidisciplinary team meeting is a clinical intervention, and we aim to ensure that any findings that generate a downstream clinical intervention are included.

**Vascular**

All the following are categorized as abnormal for vascular:

– Aneurysm

- including coiled aneurysms regardless of whether there is a residual neck or not

– Arteriovenous malformation

– Arteriovenous dural fistula

– Cavernoma

– Capillary telangiectasia

– Chronic / non-specific microhaemorrhages

– Petechial haemorrhage

– Developmental venous anomaly

– Venous sinus thrombosis

– Vasculitis if associated with vessel changes such as luminal stenosis or vessel wall enhancement

– Arterial occlusion / flow void abnormality or absence

– Venous sinus tumour invasion (this is labelled as both 'vascular' and 'mass' abnormalities)

– Arterial stenosis. If constitutional / normal variant not included.

− Vascular-like findings which are considered normal include descriptions of sluggish flow, flow-related signal abnormalities (unless they raise the suspicion of thrombus) and vascular fenestrations.

Note that findings that typically may have minimal clinical relevance when confirmed by a neuroradiology expert, are included in this category e.g., developmental venous anomaly. The rationale is that such a finding might generate a referral to a multidisciplinary team meeting for clarification of clinical relevance. We consider that a referral to a multidisciplinary team meeting is a clinical intervention and we aim to ensure that any findings that generate a downstream clinical intervention are included.

**Encephalomalacia**

All the following are categorized as abnormal for encephalomalacia:

− Gliosis

− Encephalomalacia

− Cavity

− Post-operative tissue changes / appearances are included as encephalomalacia

− Tumour debulking or partial resection as this implies residual tumour (note: these are labelled as both 'encephalomalacia' and 'mass'' abnormalities)

− Chronic infarct / sequelae of infarct

− Chronic haemorrhage / sequelae of haemorrhage (with / without haemosiderin staining)

− Cortical laminar necrosis

Encephalomalacia-like findings which are considered normal unless there is a clear description of related parenchymal injury include craniotomy, burr-holes, posterior fossa decompression, and 3rd ventriculostomy

**Acute stroke**

All the following are categorized as abnormal for acute stroke:

− Acute / subacute infarct (if demonstrating restricted diffusion)
  • Include if there are other descriptors indicating a subacute nature such as swelling even though restricted diffusion has normalised

– If a single ischaemic event with both diffusion restricting and non-restricting elements then this is labelled as an 'acute stroke' abnormality (rather than an 'encephalomalacia' abnormality)

– Parenchymal post-operative restricted diffusion secondary to retraction injury

– Mitochondrial Encephalopathy with Lactic Acidosis and Stroke-like episodes (MELAS) if associated with restricted diffusion

– Hypoxic ischaemic injury if associated with restricted diffusion

– Vasculitis if associated with acute / subacute infarct

– 'Mature', 'established', 'chronic' or 'old' infarcts without other descriptors are labelled as 'encephalomalacia' abnormalities

**White matter inflammation**

All the following are categorized as abnormal for white matter inflammation:

– Multiple sclerosis (MS) including when some plaques show cavitation (low $T_1$ signal)

– Other demyelinating lesions including Acute Disseminated Encephalomyelitis (ADEM) and Neuromyelitis Optica spectrum disorder (NMO)

– Inflammatory lesions in Radiologically Isolated Syndrome / Clinically Isolated Syndrome

– Focal cortical thinning i.e., secondary to chronic subcortical / cortical lesions, are labelled as 'encephalomalacia' abnormalities

– Progressive Multifocal Leukoencephalopathy (PML)/ Immune Reconstitution Inflammatory Syndrome (IRIS)

– Leukoencephalopathies - congenital or acquired (including toxic)

– Encephalitis / encephalopathy if it involves the white matter, e.g. related to human immunodeficiency virus (HIV) and congenital cytomegalovirus (CMV)

– Posterior Reversible Encephalopathy Syndrome (PRES)

– Osmotic demyelination (central pontine myelinolysis/ extrapontine myelinolysis)

– Susac syndrome

– Radiation if describing white matter abnormality

– White matter changes in the context of vasculitis if clearly attributed to vasculitis.

– Amyloid-related inflammatory change / inflammatory

**Atrophy**

Volume loss in excess of age

**General abnormality category:**

In addition to these 7 specialised categories, there is a generalised 'abnormal' category. This includes reports describing any abnormality from the 7 granular categories, as well as any of the following:

Hydrocephalus:

 – Acute

 – Trapped ventricle

– Chronic / stable / improving hydrocephalus (it does not matter whether its compensated or not)

 – Ventricular enlargement

 – normal pressure hydrocephalus (NPH)

Haemorrhage:

 – Any acute / subacute haemorrhage parenchymal, subarachnoid, subdural, extradural

– Acute microhaemorrhages / petechial haemorrhages

Foreign body:

– Shunts

– Clips

– Coils

– If significant metalwork is involved in skull repair e.g. in a cranioplasty (or the occasional craniotomy causing extreme intracranial MRI signal distortion)

– If craniotomies are not causing anything other than slight artefact, then these are considered normal

Extracranial:

– Total mastoid opacification / middle ear effusions

– Complete opacification / obstruction of the paranasal sinuses

– Mucosal thickening is not included

– If there is clearly a well-defined unambiguous polyp then label as abnormal.

– If 'retention cysts' or 'polypoid mucosal thickening' then label as normal. If it is something indistinguishable which could be a retention cyst / polyp then label as normal.

– Anything leading to sinus obstruction always label as abnormal.

– Calvarial / extra-calvarial masses

– Osteo-dural defects

– Encephaloceles

– Pseudomeningoceles

– Extracranial vascular abnormalities i.e., below the petrous segment e.g. cervical internal carotid artery (ICA) dissection

– Extracranial masses including lipoma or sebaceous cyst

– Orbital abnormalities

- Including optic nerve pathology affecting the orbital segment of the nerve i.e., meningioma

– Cases with isolated tortuous optic nerve sheath complexes with no other features suggestive of raised intracranial pressure, are labelled as normal

– Eye prostheses and proptosis

– Pseudophakia is labelled as normal

– Bone abnormality e.g., low bone signal secondary to haemoglobinopathy

– Basilar invagination

– Hyperostosis is considered normal

– Thornwald cysts are considered normal

Intracranial miscellaneous:

– Cerebellar ectopia

– Brain herniation (e.g., through a craniectomy defect)

– Clear evidence of intracranial hypertension (e.g., prominent optic nerve sheaths AND intrasellar subarchnoid herniation)

- Isolated intrasellar subarachnoid herniation / empty sella is labelled normal

- Isolated tapering of dural venous sinuses is labelled normal

– Clear evidence of intracranial hypotension (e.g., pituitary enlargement AND pachymeningeal thickening)

- If subdural collections present, these are also labelled as 'mass'

– Cerebral oedema or reduced CSF spaces from parenchymal swelling

– Absent or hypoplastic structures such as agenesis of the corpus callosum

– Meningeal thickening or enhancement for example in the context of neurosarcoid or vasculitis

– Enhancing or thickened cranial nerves

– Infective processes primarily involving the meninges or ependyma (i.e. ventriculitis or meningitis)

– Encephalitis if primarily involving the cortex (herpes simplex virus (HSV)/ autoimmune encephalitis)

– Excessive or unexpected basal ganglia or parenchymal calcification

– Optic neuritis involving the intracranial segments of the optic nerves or chiasmitis

– Adhesions / webs

– Pneumocephalus

– Colpocephaly

– Superficial siderosis

– Ulegyria

– Focal areas of signal intensity (FASIs) / Unidentified bright objects (UBO)

– Basal ganglia / thalamic changes in the context of metabolic abnormalities

– Neurovascular conflict fulfilling conditions of nerve distortion AND nerve root entry zone involvement

– Band heterotopia and polymicrogyria

– Hypophysitis

– Seizure related changes

– Amyotrophic lateral sclerosis (ALS).

**Appendix D**

Examples of true and false Findings and Summary pairs used for our self-supervised radiology section matching pre-training task.

1. **True pairs**

-------------------------------------------------------------------------------------------------------------------

**Findings:** There is a small focus of restricted diffusion involving the posterior aspect of the left corona radiata in keeping with a subacute infarct. There are patchy areas of $T_2$ hyperintensity involving the subcortical/deep white matter, pons and left thalamus. There is more confluent smaller $T_2$ hyperintense foci within the basal ganglia and sub insular regions bilaterally representing a combination of small lacunar infarcts and prominent peri vascular spaces. Overall, the appearances are in keeping with moderate small vessel ischaemic change.

**Summary:** Small subacute infarct involving the posterior aspect of the left corona radiata and features to suggest moderately severe small vessel ischaemic change

-------------------------------------------------------------------------------------------------------------------

**Findings:** There is a mass lesion of heterogenous signal within the right cerebellar hemisphere. This extends inferiorly through the foramen of Luschka with an exophytic component that compresses and displaces the medulla anteriorly and to the left. It extends superomedially to the roof and right lateral margin of the fourth ventricle. There is infiltration of the inferior right cerebellar hemisphere and the right middle cerebellar peduncles. The mass enhances heterogeneously and demonstrates peripheral partial restriction of diffusion. There is a central $T_2$ hyperintense non-enhancing component with free diffusion. The third and lateral ventricles are not enlarged. The supratentorial appearances are normal. There is no pathological enhancement within the spinal canal.

**Summary:** The imaging features are those of a high-grade right cerebellar tumour, likely ependymoma

-----------------------------------------------------------------------------------------------------------

**Findings:** There are minor generalised changes of involution with ventricular and sulcal prominence. The degree of minor and generalised volume loss is commensurate with the patients age. There is diffuse high signal on $T_2$/FLAIR within the white matter of both cerebral hemispheres.  this is predominantly periventricular with extension into the subcortical white matter of both frontal and parietal lobes. There is sparing of the corpus collosum and subcortical U-fibres.

**Summary:** The imaging features are those of severe small vessel ischaemic change within the white matter of both cerebral hemispheres

-----------------------------------------------------------------------------------------------------------

**Findings:** There are generalised changes of involution with ventricular and sulcal prominence. Although this volume loss is generalised there is a predominant temporal volume loss with bilateral hippocampal atrophy and reduction in volume of entorhinal cortex. The changes are asymmetrical with more marked frontal and temporal volume loss on the right. No focal abnormalities of the brain parenchyma have been identified.

**Summary:** The imaging features are of those of an Alzheimer pattern of volume loss predominantly affecting the temporal lobes. These changes are asymmetrical and more marked on the right

   2. **False pairs**

-----------------------------------------------------------------------------------------------------------

**Findings:** There is a large, heterogenous lesion in the right frontal lobe with necrotic components and extensive surrounding vasogenic oedema with mass effect, sulcal effacement and partial effacement of right lateral ventricle. There is approximately 11 mm of contralateral midline shift to the left. A smaller lesion is seen at the grey-white matter junction in the left precentral gyrus and this also has significant surrounding vasogenic oedema. These lesions are associated with abnormal susceptibility artefact in keeping with blood degradation products and they demonstrate heterogenous, mainly peripheral, enhancement. Smaller enhancing lesions are seen in the cerebellar hemispheres bilaterally. The post-contrast sequences also demonstrate small enhancing metastases in the left frontal, left medial temporal and in the left occipital lobes. Prominent linear subarachnoid enhancement is presumed all vascular. None of the lesions demonstrate any definite restricted diffusion. No bony destructive lesion is identified.

**Summary:** Normal intracranial appearances.

---

**Findings:** No acute intracranial abnormality is demonstrated. Specifically, there is no acute infarct. No microhaemorrhages have occurred. The major intracranial vessels return their normal flow related signal voids. The intracranial appearances are unremarkable.

**Summary:** There is a large right intra axial heterogenous enhancing mass lesion consistent with a high-grade glioma. There is associated midline shift with dilatation of the left lateral ventricle.

---

**Findings:** There are multiple partially confluent foci, hyperintense on the T2 sequences, with no restriction on the DWI, located in the deep and subcortical frontoparietal white matter of both frontal cerebral hemispheres, with no evidence of involvement of the U-fibres. The cerebellum, brainstem and corpus callosum are unremarkable. The described findings are more likely to be of vascular or residual origin than of demyelinating (multiple sclerosis type) origin. The ventricles and cortical sulci are of normal size and appearance. Normal cranio-cervical junction. There is no fluid in the mastoid air cells

**Summary:** Solitary 5mm cystic lesion in the left para-hippocampal gyrus. This probably represents a small neuroglial cyst.

---

**Findings:** The cranio-cervical junction is normal. The brainstem and cerebellum are unremarkable. The cerebral hemispheres are normal in morphology and signal intensity pattern. There is no evidence of intracranial ischaemic or haemorrhagic lesions. There is no evidence of intracranial solid or cystic lesions. The ventricular system is normal in size.

**Summary:** The appearances are those of longstanding compensated communicating hydrocephalus.

-------------------------------------------------------------------------------------------------------------------

**Appendix E**

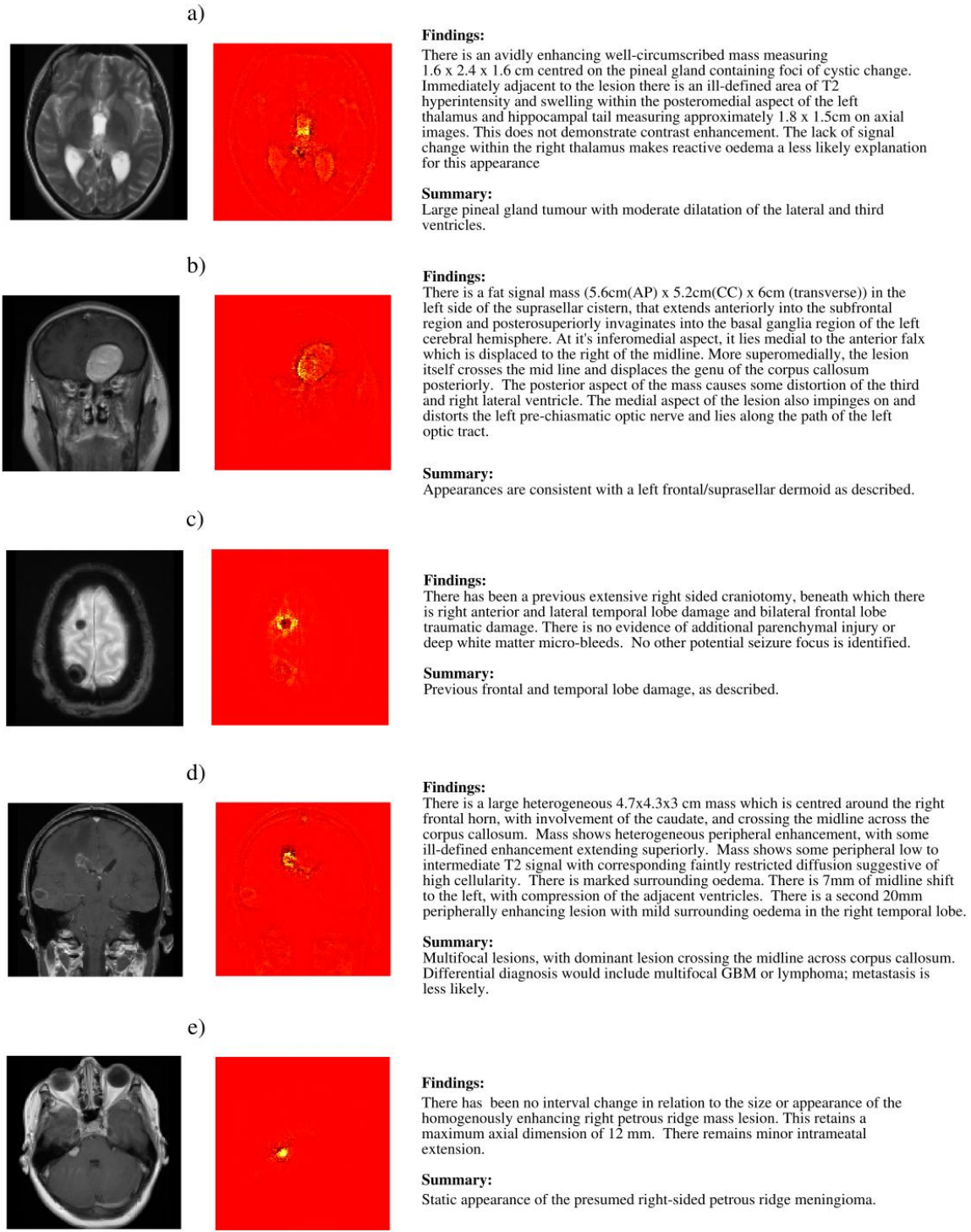

**Figure E1:** *Examples where our text-vision model retrieved images which do not contain to the pathology requested; in each case, ambiguous or morphologically similar pathologies were returned. Heatmaps generated by guided backpropagation are also shown, along with the corresponding radiology report for each examination (which was not used for image retrieval). a) retrieval task: find image examples of a pineal cyst; incorrectly retrieved example: the image shows a pineal gland; b) retrieval task: find image examples of*

*metastatic deposits; incorrectly retrieved example: the image shows a dermoid, which has similar imaging characteristics to a meningioma; c) retrieval task: find image examples of a haematoma; incorrectly retrieved example: an image showing susceptibility artefacts on GRE due to bolts holding the patient's craniotomy together which has been mistaken for susceptibility resulting from a haematoma; d) retrieval task: find image examples of metastases; incorrectly retrieved example: the image shows a presumed glioblastoma or lymphoma (although metastasis is in the differential diagnosis); e) retrieval task: find image examples of a vestibular schwannoma; incorrectly retrieved example: the image shows a right-sided petrous ridge meningioma (although in this location, vestibular schwannoma is in the differential diagnosis). Note: using radiological left and right.*

**Appendix F**

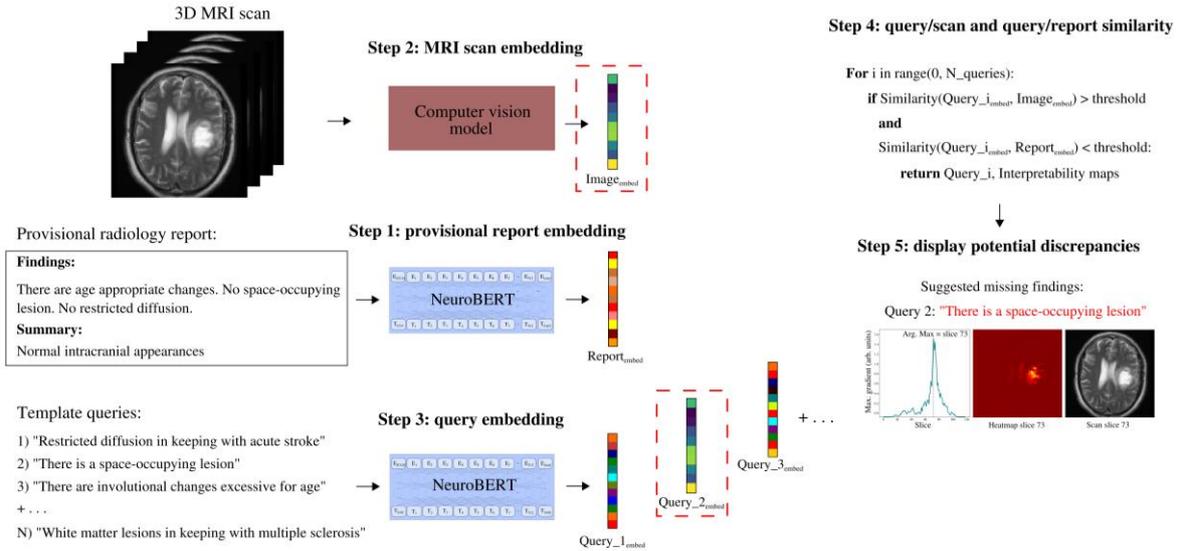

**Figure F1:** *Overview of our proposed approach to detect errors in provisional radiology reports. Before finalizing a radiologist's provisional report, it could undergo an additional validation step through our framework. Step 1) the report would be encoded into a 768-dimensional embedding by NeuroBERT. Step 2) the MRI scans from the examination would be similarly processed into embeddings by the corresponding single-sequence computer vision models. Step 3) A series of template queries, representing key pathological findings (such as 'acute stroke' or 'space occupying lesion'), would be encoded into the same vector space. Step 4) the framework would then calculate the cosine similarity between the query embeddings and both the scan embeddings and the provisional report embedding. Step 5) discrepancies highlighted by high similarity with the scan but low similarity with the provisional report could signal potential oversights. These flagged queries, along with their heatmaps, would be presented to radiologists for a focused review of the implicated slices.*